\documentclass[runningheads]{llncs}

\usepackage{graphicx}
\usepackage{tikz}
\usepackage{comment}
\usepackage{microtype}
\usepackage{amsmath,amssymb,amsfonts,dsfont,bm,mathrsfs}
\usepackage{booktabs,multirow,adjustbox}
\usepackage{color}
\usepackage{url}
\usepackage[colorlinks,linkcolor=blue]{hyperref}
\makeatletter
\newcommand{\thankssymbol}[1]{\textsuperscript{\@fnsymbol{#1}}}
\makeatother

\begin{document}
\pagestyle{headings}
\mainmatter
\def\ECCVSubNumber{2839}

\newcommand{\E}{\mathbb{E}}                   
\newcommand{\z}{{\rm\bf z}}                   
\newcommand{\w}{{\rm\bf w}}                   
\newcommand{\Z}{\mathcal{Z}}                  
\newcommand{\W}{\mathcal{W}}                  
\renewcommand{\S}{\mathcal{S}}                
\newcommand{\x}{{\rm\bf x}}                   
\newcommand{\X}{\mathcal{X}}                  
\newcommand{\D}{\mathcal{D}}                  
\newcommand{\n}{{\rm\bf n}}                   
\newcommand{\Loss}{\mathcal{L}}               
\newcommand{\FID}{\textbf{FID$\downarrow$}}  
\newcommand{\SWD}{\textbf{SWD$\downarrow$}}  
\newcommand{\MSE}{\textbf{MSE$\downarrow$}}  

\def\httilde{\mbox{\tt\raisebox{-.5ex}{\symbol{126}}}}

\title{In-Domain GAN Inversion\\for Real Image Editing}
\titlerunning{In-Domain GAN Inversion}
\author{
  Jiapeng Zhu\thanks{denotes equal contribution.}\inst{1}\and
  Yujun Shen\thankssymbol{1}\inst{1}\and
  Deli Zhao\inst{2}\and
  Bolei Zhou\inst{1}
}
\authorrunning{J. Zhu\thankssymbol{1}, Y. Shen\thankssymbol{1}, D. Zhao, B. Zhou}
\institute{
  The Chinese University of Hong Kong\\
  \email{\{jpzhu, sy116, bzhou\}@ie.cuhk.edu.hk}\\ \and
  Xiaomi AI Lab\\
  \email{zhaodeli@gmail.com}
}

\maketitle

\begin{abstract}
  Recent work has shown that a variety of semantics emerge in the latent space of Generative Adversarial Networks (GANs) when being trained to synthesize images.
  However, it is difficult to use these learned semantics for real image editing.
  A common practice of feeding a real image to a trained GAN generator is to invert it back to a latent code.
  However, existing inversion methods typically focus on reconstructing the target image by pixel values yet fail to land the inverted code in the semantic domain of the original latent space.
  As a result, the reconstructed image cannot well support semantic editing through varying the inverted code.
  To solve this problem, we propose an \emph{in-domain} GAN inversion approach, which not only faithfully reconstructs the input image but also ensures the inverted code to be semantically meaningful for editing.
  We first learn a novel \emph{domain-guided} encoder to project a given image to the native latent space of GANs.
  We then propose \emph{domain-regularized} optimization by involving the encoder as a regularizer to fine-tune the code produced by the encoder and better recover the target image.
  Extensive experiments suggest that our inversion method achieves satisfying real image reconstruction and more importantly facilitates various image editing tasks, significantly outperforming start-of-the-arts.\footnote[1]{Code and models are available at \url{https://genforce.github.io/idinvert/}.}
  %
\end{abstract}

\begin{figure}[t]
  \centering
  \includegraphics[width=1.0\linewidth]{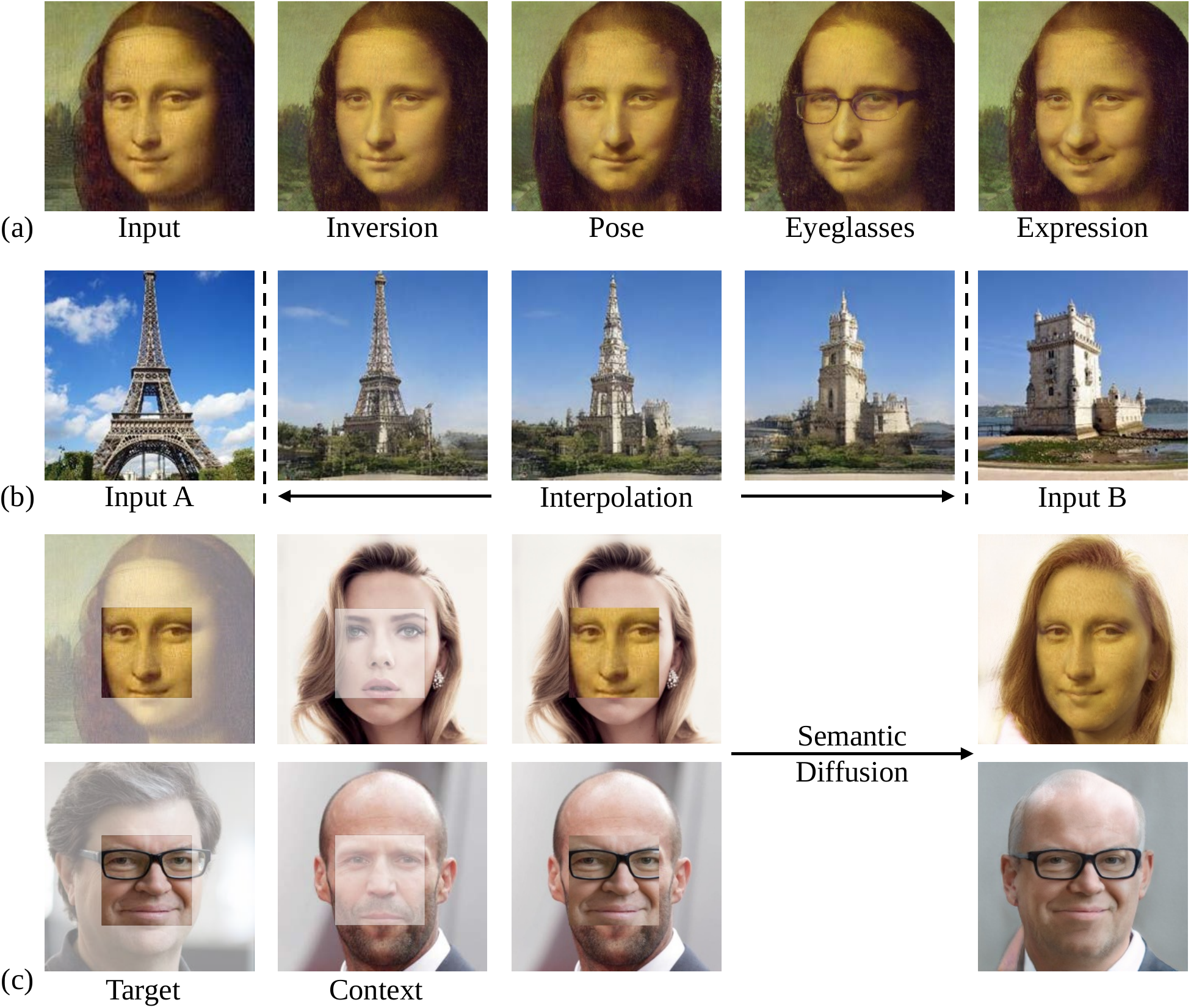}
  \caption{
    Real image editing using the proposed \emph{in-domain} GAN inversion with a \emph{fixed} GAN generator.
    (a) Semantic manipulation with respect to various facial attributes.
    (b) Image interpolation by linearly interpolating two inverted codes.
    (c) Semantic diffusion which diffuses the target face to the context and makes them compatible.
  }
  \label{fig:teaser}
\end{figure}

\section{Introduction}
Generative Adversarial Networks (GANs) \cite{gan} are formulated as a two-player game between a generator to synthesize images and a discriminator to differentiate real data from fake data.
Recent work \cite{goetschalckx2019ganalyze,gansteerability,shen2019interpreting} has shown that GANs spontaneously learn to encode rich semantics inside the latent space and that varying the latent code leads to the manipulation of the corresponding attributes occurring in the output image.
However, it remains difficult to apply such manipulation capability to real images since GANs lack the ability of taking a particular image as the input to infer its latent code.

Many attempts have been made to reverse the generation process by mapping the image space back to the latent space, which is widely known as \emph{GAN inversion}.
They either learn an extra encoder beyond the GAN \cite{luo2017learning,zhu2016generative,bau2019inverting} or directly optimize the latent code for an individual image \cite{lipton2017precise,invertibility,creswell2018inverting}.
However, existing methods mainly focus on reconstructing the pixel values of the input image, leaving some important open questions about the property of the inverted code.
For example, does the inverted code lie in the original latent space of GANs?
Can the inverted code semantically represent the target image?
Does the inverted code support image editing by reusing the knowledge learned by GANs?
Can we use a well-trained GAN to invert any image?
Answering these questions not only deepens our understanding of the internal mechanism of GANs, but is also able to unleash the pre-trained GAN models for the versatile image editing capability.

In this work, we show that a good GAN inversion method should not only reconstruct the target image at the \emph{pixel} level, but also align the inverted code with the \emph{semantic} knowledge encoded in the latent space.
We call such semantically meaningful codes as \emph{in-domain} codes since they are subject to the semantic domain learned by GANs.
We also find that in-domain codes can better support image editing by reusing the rich knowledge emerging in the GAN models.
To this end, we propose an \emph{in-domain} GAN inversion approach to recover the input image at \emph{both the pixel level and the semantic level}.
Concretely, we first train a novel \emph{domain-guided} encoder to map the image space to the latent space such that all codes produced by the encoder are in-domain.
We then perform instance-level \emph{domain-regularized} optimization by involving the encoder as a regularizer to better reconstruct the pixel values without affecting the semantic property of the inverted code.
We summarize our contributions as follows:
\begin{itemize}
  \item We analyze an important issue in the GAN inversion task that the inverted code should go beyond merely recovering the per-pixel values of the input image by further considering the semantic information.
  \item We propose an \emph{in-domain} GAN inversion approach by first learning a \emph{domain-guided} encoder and further use this encoder as a regularizer for \emph{domain-regularized} optimization.
  \item We evaluate our method on a variety of image editing tasks, as shown in Fig.\ref{fig:teaser}. Qualitative and Quantitative results suggest that our \emph{in-domain} inversion can faithfully recover the target image from both the low-level pixels and the high-level semantics, significantly surpassing existing approaches.
\end{itemize}

\subsection{Related Work}\label{subsec:related-work}

\noindent\textbf{Generative Adversarial Networks (GANs).}
By learning the distribution of real images via adversarial training, GANs \cite{gan} have advanced image synthesis in recent years.
Many variants of GANs are proposed to improve the synthesis quality \cite{sngan,sagan,pggan,biggan,stylegan} and training stability \cite{wgan,wgan_gp,began}.
Recently, GANs are shown to spontaneously learn semantics inside the latent space, which can be further used to control the generation process.
Goetschalckx \emph{et al.} \cite{goetschalckx2019ganalyze} explored how to make the synthesis from GANs more memorable, Jahanian \emph{et al.} \cite{gansteerability} achieved camera movements and color changes by shifting the latent distribution, Shen \emph{et al.} \cite{shen2019interpreting} interpreted the latent space of GANs for semantic face editing, and Yang \emph{et al.} \cite{yang2019semantic} observed that hierarchical semantics emerge from the layer-wise latent codes of GANs for scene synthesis.
However, due to the lack of inference capability in GANs, it remains difficult to apply the rich semantics encoded in the latent space to editing real images.

\noindent\textbf{GAN Inversion.}
To better apply well-trained GANs to real-world applications, GAN inversion enables real image editing from the latent space \cite{zhu2016generative,perarnau2016invertible,bau2019semantic}.
Given a fixed GAN model, GAN inversion aims at finding the most accurate latent code to recover the input image.
Existing inversion approaches typically fall into two types.
One is learning-based, which first synthesizes a collection of images with randomly sampled latent codes and then uses the images and codes as inputs and supervisions respectively to train a deterministic model \cite{perarnau2016invertible,zhu2016generative}.
The other is optimization-based, which deals with a single instance at one time by directly optimizing the latent code to minimize the pixel-wise reconstruction loss \cite{lipton2017precise,creswell2018inverting,invertibility,image2stylegan}.
Some work combines these two ideas by using the encoder to generate an initialization for optimization \cite{bau2019seeing,bau2019inverting}.
There are also some models that take invertibility into account at the training stage by designing new architectures \cite{ali,bigan,lia,glow}.
Some concurrent work improves GAN inversion with better reconstruction quality:
Gu \emph{et al.} \cite{gu2020image} employs multiple latent codes to recover a single image, Pan \emph{et al.} \cite{pan2020exploiting} optimizes the parameters of the generator together with the latent code, Karras \emph{et al.} \cite{stylegan2} and Abdal \emph{et al.} \cite{image2stylegan++} focus on inverting StyleGAN \cite{stylegan} models by exploiting the layer-wise noises.

\noindent\textbf{Key Difference.}
One important issue omitted by existing inversion methods is that they merely focus on reconstructing the target image at the pixel level without considering the semantic information in the inverted code.
If the code cannot align with the semantic domain of the latent space, even being able to recover the per-pixel values of the input image, it would still fail to reuse the knowledge learned by GANs for semantic editing.
Therefore, in this work, we argue that only using the pixel-wise reconstruction loss as the metric to evaluate a GAN inversion approach is not proper enough.
Instead, we deeply study the property of the inverted code from the \emph{semantic} level and propose the \emph{in-domain} GAN inversion that well supports real image editing.

\section{In-Domain GAN Inversion}\label{sec:method}
As discussed above, when inverting a GAN model, besides recovering the input image by pixel values, we also care about whether the inverted code is semantically meaningful.
Here, the semantics refer to the emergent knowledge that GAN has learned from the observed data \cite{goetschalckx2019ganalyze,gansteerability,shen2019interpreting,yang2019semantic}.
For this purpose, we propose to first train a \emph{domain-guided} encoder and then use this encoder as a regularizer for the further \emph{domain-regularized} optimization, as shown in Fig.\ref{fig:framework}.

\definecolor{royalazure}{rgb}{0.0, 0.22, 0.66}
\begin{figure}[t]
  \centering
  \includegraphics[width=1.0\linewidth]{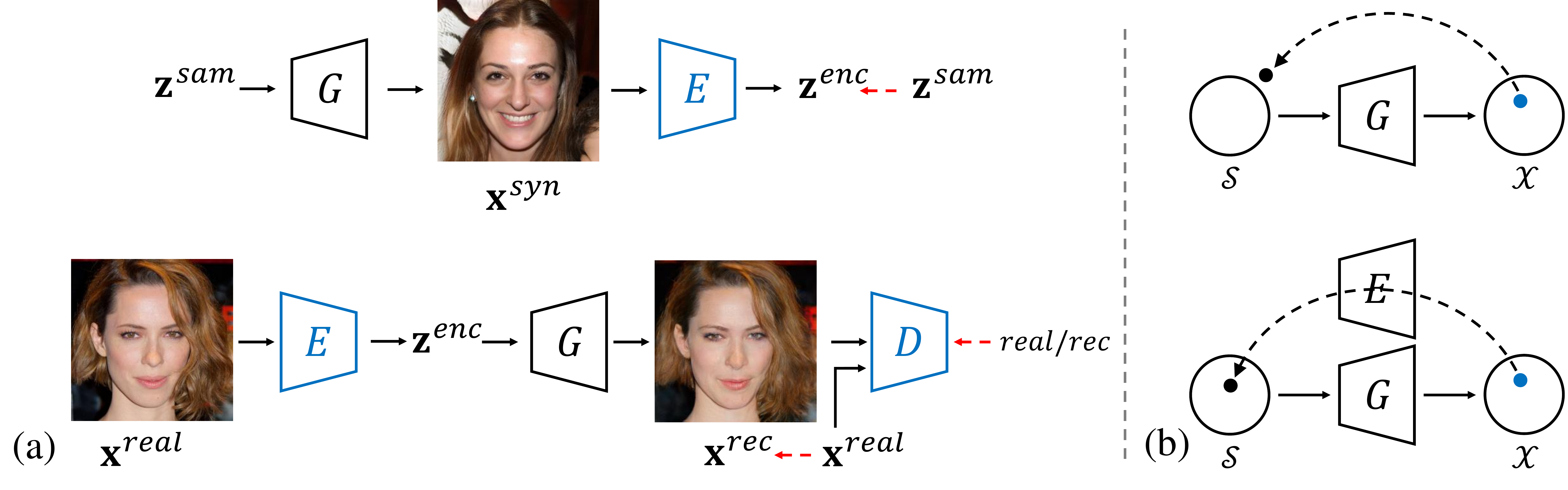}
  \caption{
    (a) The comparison between the training of conventional encoder and \emph{domain-guided} encoder for GAN inversion.
    Model blocks in \textbf{\textcolor{royalazure}{blue}} are trainable and \textbf{\textcolor{red}{red}} dashed arrows indicate the supervisions.
    Instead of being trained with synthesized data to recover the latent code, our \emph{domain-guided} encoder is trained with the objective to recover the real images.
    The \emph{fixed} generator is involved to make sure the codes produced by the encoder lie in the native latent space of the generator and stay semantically meaningful.
    (b) The comparison between the conventional optimization and our \emph{domain-regularized} optimization.
    The well-trained \emph{domain-guided} encoder is included as a regularizer to land the latent code in the semantic domain during the optimization process.
  }
  \label{fig:framework}
\end{figure}

\noindent\textbf{Problem Statement.}
Before going into details, we briefly introduce the problem setting with some basic notations.
A GAN model typically consists of a generator $G(\cdot): \Z\rightarrow\X$ to synthesize high-quality images and a discriminator $D(\cdot)$ to distinguish real from synthesized data.
GAN inversion studies the reverse mapping of $G(\cdot)$, which is to find the best latent code $\z^{inv}$ to recover a given real image $\x^{real}$.
We denote the semantic space learned by GANs as $\S$.
We would like $\z^{inv}$ to also align with the prior knowledge $\S$ in the pre-trained GAN model.

\noindent\textbf{Choice of Latent Space.}
Typically, GANs sample latent codes $\z$ from a pre-defined distributed space $\Z$, such as normal distribution.
The recent StyleGAN model \cite{stylegan} proposes to first map the initial latent space $\Z$ to a second latent space $\W$ with Multi-Layer Perceptron (MLP), and then feed the codes $\w\in\W$ to the generator for image synthesis.
Such additional mapping has already been proven to learn more disentangled semantics \cite{stylegan,shen2019interpreting}.
As a result, the disentangled space $\W$ is widely used for the GAN inversion task \cite{image2stylegan,lia,image2stylegan++,stylegan2}.
Similarly, we also choose $\W$ space as the inversion space for three reasons:
(i) We focus on the semantic (\emph{i.e.,} \emph{in-domain}) property of the inverted codes, making $\W$ space more appropriate for analysis.
(ii) Inverting to $\W$ space achieves better performance than $\Z$ space \cite{lia}.
(iii) It is easy to introduce the $\W$ space to any GAN model by simply learning an extra MLP ahead of the generator. Hence, it will not harm the generalization ability of our approach.
In this work, we conduct all experiments on the $\W$ space, but our approach can be performed on the $\Z$ space as well.
For simplicity, we use $\z$ to denote the latent code in the following sections.

\subsection{Domain-Guided Encoder}\label{subsec:domain-guided-encoder}
Training an encoder is commonly used for GAN inversion problem \cite{perarnau2016invertible,zhu2016generative,bau2019seeing,bau2019inverting} considering its fast inference speed.
However, existing methods simply learn a deterministic model with no regard to whether the codes produced by the encoder align with the semantic knowledge learned by $G(\cdot)$.
As shown on the top of Fig.\ref{fig:framework}(a), a collection of latent codes $\z^{sam}$ are randomly sampled and fed into $G(\cdot)$ to get the corresponding synthesis $\x^{syn}$.
Then, the encoder $E(\cdot)$ takes $\x^{syn}$ and $\z^{sam}$ as inputs and supervisions respectively and is trained with
\begin{align}
  \min_{\Theta_E}\Loss_E = ||\z^{sam} - E(G(\z^{sam}))||_2, \label{eq:conventional-encoder}
\end{align}
where $||\cdot||_2$ denotes the $l_2$ distance and $\Theta_E$ represents the parameters of the encoder $E(\cdot)$.
We argue that the supervision by only reconstructing $\z^{sam}$ is not powerful enough to train an accurate encoder.
Also, the generator is actually omitted and cannot provide its domain knowledge to guide the training of encoder since the gradients from $G(\cdot)$ are not taken into account at all.

To solve these problems, we propose to train a \emph{domain-guided} encoder, which is illustrated in the bottom row of Fig.\ref{fig:framework}(a).
There are three main \textbf{differences} compared to the conventional encoder:
(i) The output of the encoder is fed into the generator to reconstruct the input image such that the objective function comes from the image space instead of latent space. This involves semantic knowledge from the generator in training and provides more informative and accurate supervision. The output code is therefore guaranteed to align with the semantic domain of the generator.
(ii) Instead of being trained with synthesized images, the \emph{domain-guided} encoder is trained with real images, making our encoder more applicable to real applications.
(iii) To make sure the reconstructed image is realistic enough, we employ the discriminator to compete with the encoder. In this way, we can acquire as much information as possible from the GAN model (\emph{i.e.}, both two components of GAN are used). The adversarial training manner also pushes the output code to better fit the semantic knowledge of the generator.
We also introduce perceptual loss \cite{johnson2016perceptual} by using the feature extracted by VGG \cite{vgg}.
Hence, the training process can be formulated as
\begin{align}
  &\begin{aligned}
    \min_{\Theta_E}\Loss_E =  ||\x^{real} - G(E(\x^{real}))||_2\
                              &+\lambda_{vgg} ||F(\x^{real}) - F(G(E(\x^{real})))||_2\ \\
                              &-\lambda_{adv} \underset{\x^{real}\sim P_{data}}\E[D(G(E(\x^{real})))], \label{eq:encoder}
  \end{aligned} \\
  &\begin{aligned}
    \min_{\Theta_D}\Loss_D =  \underset{\x^{real}\sim P_{data}}\E[D(G(E(\x^{real})))]\
                              &-\underset{{\x^{real}\sim P_{data}}}\E[D(\x^{real})]\ \\
                              &+\frac{\gamma}{2}\underset{{\x^{real}\sim P_{data}} }{\E}[||\nabla_{{\x}}D(\x^{real})||_2^2],
  \end{aligned} \label{eq:discriminator}
\end{align}
where $P_{data}$ denotes the distribution of real data and $\gamma$ is the hyper-parameter for the gradient regularization.
$\lambda_{vgg}$ and $\lambda_{adv}$ are the perceptual and discriminator loss weights.
$F(\cdot)$ denotes the VGG feature extraction model.

\subsection{Domain-Regularized Optimization}\label{subsec:domain-regularized-optimization}
Unlike the generation process of GANs which learns a mapping at the distribution level, \emph{i.e.} from latent distribution to real image distribution, GAN inversion is more like an instance-level task which is to best reconstruct a given individual image.
From this point of view, it is hard to learn a perfect reverse mapping with an encoder alone due to its limited representation capability.
Therefore, even though the inverted code from the proposed \emph{domain-guided} encoder can well reconstruct the input image based on the pre-trained generator and ensure the code itself to be semantically meaningful, we still need to refine the code to make it better fit the target individual image at the pixel values.

Previous methods \cite{creswell2018inverting,invertibility,image2stylegan} propose to gradient descent algorithm to optimize the code.
The top row of Fig.\ref{fig:framework}(b) illustrates the optimization process where the latent code is optimized ``freely'' based on the generator only.
It may very likely produce an out-of-domain inversion since there are no constraints on the latent code at all.
Relying on our \emph{domain-guided} encoder, we design a \emph{domain-regularized} optimization with two improvements, as shown at the bottom of Fig.\ref{fig:framework}(b):
(i) We use the output of the \emph{domain-guided} encoder as an ideal starting point which avoids the code from getting stuck at a local minimum and also significantly shortens the optimization process.
(ii) We include the \emph{domain-guided} encoder as a regularizer to preserve the latent code within the semantic domain of the generator.
To summarize, the objective function for optimization is
\begin{align}
  \begin{aligned}
    \z^{inv} = \arg\min_{\z}\ ||\x - G(\z)||_2\ &+ \lambda_{vgg}||F(\x) - F(G(\z))||_2\ \\
                                                &+ \lambda_{dom}||\z - E(G(\z))||_2,
  \end{aligned} \label{eq:optimization}
\end{align}
where $\x$ is the target image to invert.
$\lambda_{vgg}$ and $\lambda_{dom}$ are the loss weights corresponding to the perceptual loss and the encoder regularizer respectively.

\section{Experiments}\label{sec:experiments}
In this section, we experimentally show the superiority of the proposed \emph{in-domain} GAN inversion over existing methods in terms of semantic information preservation, inversion quality, inference speed, as well as real image editing.

\subsection{Experimental Settings}
We conduct experiments on FFHQ dataset \cite{stylegan}, which contains 70,000 high-quality face images, and LSUN dataset \cite{lsun}, which consists of images from 10 different scene categories.
Only results on the tower category are shown in the main paper.
More results can be found in the \textbf{Appendix}.
The GANs to invert are pre-trained following StyleGAN \cite{stylegan}.\footnote[2]{Different from StyleGAN, we use different latent codes for different layers.}
When training the encoder, the generator is \emph{fixed} and we only update the encoder and discriminator according to Eq.\eqref{eq:encoder} and Eq.\eqref{eq:discriminator}.
As for the perceptual loss in Eq.\eqref{eq:encoder}, we take $\mathtt{conv4\_3}$ as the VGG \cite{vgg} output.
Loss weights are set as $\lambda_{vgg}=5e^{-5}$, $\lambda_{adv}=0.1$, and $\gamma=10$.
We set $\lambda_{dom}=2$ in Eq.\eqref{eq:optimization} for the \emph{domain-regularized} optimization.

\subsection{Semantic Analysis of the Inverted Codes}
In this part, we evaluate how the inverted codes can semantically represent the target images.
As pointed out by prior work \cite{stylegan,shen2019interpreting}, the latent space of GANs is linearly separable in terms of semantics.
In particular, for a binary attribute (\emph{e.g.}, male \emph{v.s.} female), it is possible to find a latent hyperplane such that all points from the same side correspond to the same attribute.
We use this property to evaluate the alignment between the inverted codes and the latent semantics.

\begin{figure}[t]
  \centering
  \includegraphics[width=1.0\linewidth]{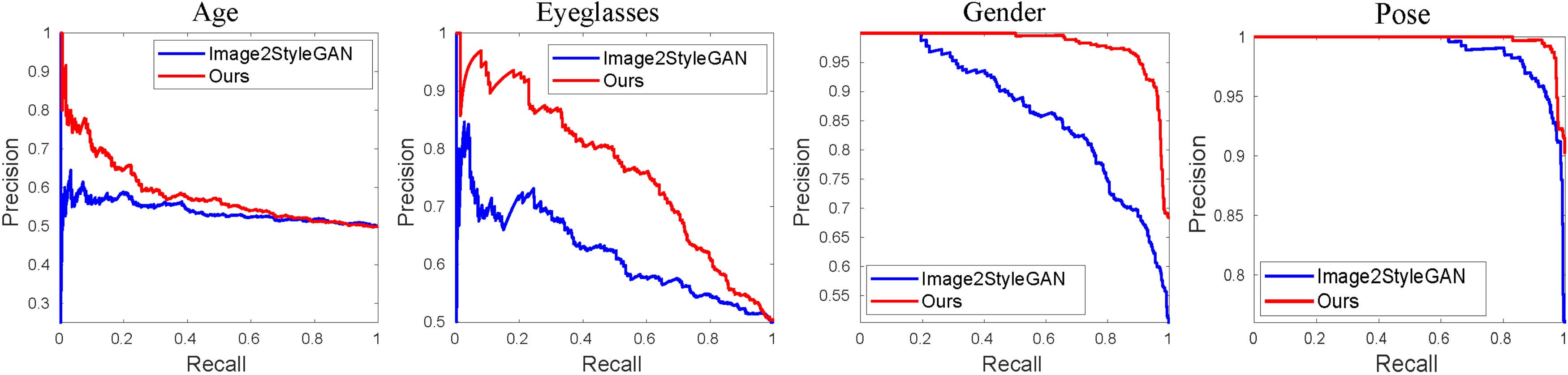}
  \caption{
    Precision-recall curves by directly using the inverted codes for facial attribute classification.
    Our \emph{in-domain} inversion shows much better performance than Image2StyleGAN \cite{image2stylegan}, suggesting a stronger semantic preservation.
  }
  \label{fig:domain}
\end{figure}

We collect 7,000 real face images and use off-the-shelf attribute classifiers to predict age (young \emph{v.s.} old), gender (female \emph{v.s.} male), eyeglasses (absence \emph{v.s.} presence), and pose (left \emph{v.s.} right).
These predictions are considered as ground-truth.
Then, we use the state-of-the-art GAN inversion method, Image2StyleGAN \cite{image2stylegan}, and our proposed \emph{in-domain} GAN inversion to invert these images back to the latent space of a \emph{fixed} StyleGAN model trained on FFHQ dataset \cite{stylegan}.
InterFaceGAN \cite{shen2019interpreting} is used to search the semantic boundaries for the aforementioned attributes in the latent space.
Then, we use these boundaries as well as the inverted codes to evaluate the attribute classification performance.
Fig.\ref{fig:domain} shows the precision-recall curves on each semantic.
We can easily tell that the codes inverted by our method are more semantically meaningful.
This quantitatively demonstrates the effectiveness of our proposed \emph{in-domain} inversion for preserving the semantics property of the inverted code.

\setlength{\tabcolsep}{4.5pt}
\begin{table}[t]
  \caption{
    Quantitative comparison between different inversion methods.
    For each model, we invert 500 images for evaluation.
    $\downarrow$ means lower number is better.
  }
  \label{tab:reconstruction_quat}
  \scriptsize\centering
  \begin{tabular}{|l|c|ccc|ccc|}
    \hline
                        \multicolumn{2}{|c|}{}                     &             \multicolumn{3}{c|}{Face}            &             \multicolumn{3}{c|}{Tower}           \\ \hline
    \textbf{Method}                              &  \textbf{Speed} &           \FID &           \SWD &           \MSE &           \FID &           \SWD &           \MSE \\ \hline
    Traditional Encoder \cite{zhu2016generative} & \textbf{0.008s} &          88.48 &          100.5 &          0.507 &          73.02 &          69.19 &          0.455 \\
    MSE-based Optimization \cite{image2stylegan} &            290s &          58.04 &          29.19 & \textbf{0.026} &          69.16 &          55.35 &          0.068 \\ \hline
    Domain-Guided Encoder (Ours)                 &          0.017s &          52.85 & \textbf{13.02} &          0.062 &          46.81 &          27.13 &          0.071 \\
    In-Domain Inversion (Ours)                   &              8s & \textbf{42.64} &          13.44 &          0.030 & \textbf{44.77} & \textbf{26.44} & \textbf{0.052} \\ \hline
  \end{tabular}
\end{table}

\begin{figure}[t]
  \centering
  \includegraphics[width=1.0\linewidth]{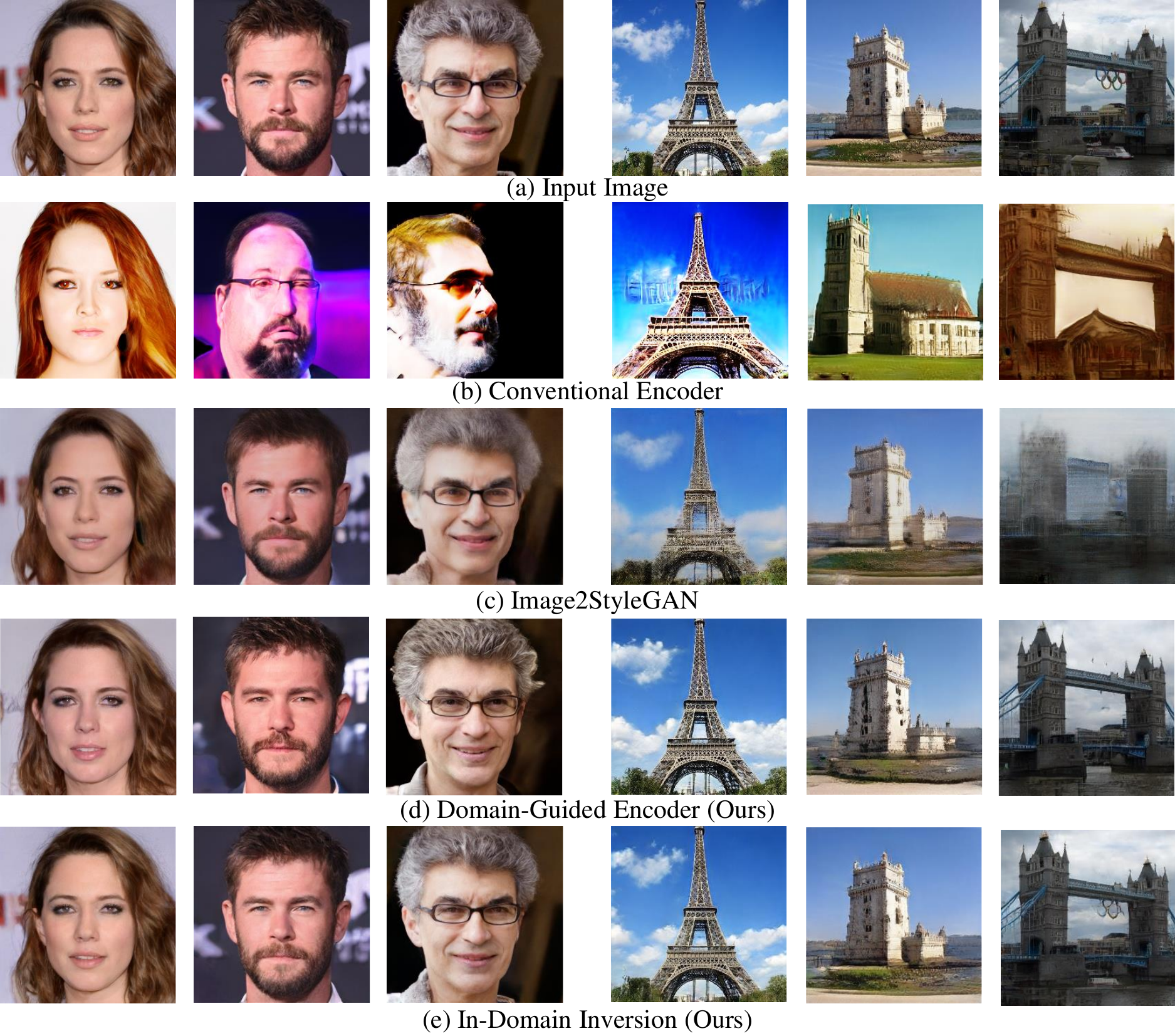}
  \caption{
    Qualitative comparison on image reconstruction with different GAN inversion methods.
    (a) Input image.
    (b) Conventional encoder \cite{zhu2016generative}.
    (c) Image2StyleGAN \cite{image2stylegan}.
    (d) Our proposed \emph{domain-guided} encoder.
    (e) Our proposed \emph{in-domain} inversion.
  }
  \label{fig:inversion}
\end{figure}

\subsection{Inversion Quality and Speed}
As discussed above, our method can produce \emph{in-domain} codes for the GAN inversion task.
In this part, we would like to verify that the improvement of our approach from the semantic aspect does not affect its performance on the traditional evaluation metric, \emph{i.e.}, image reconstruction quality.
Fig.\ref{fig:inversion} shows the qualitative comparison between different inversion methods including training traditional encoder \cite{zhu2016generative}, MSE-based optimization \cite{image2stylegan}, as well as our proposed \emph{domain-guided} encoder and the \emph{in-domain} inversion.
Comparison between Fig.\ref{fig:inversion}(b) and Fig.\ref{fig:inversion}(d) shows the superiority of our \emph{domain-guided} encoder in learning a better mapping from the image space to the latent space.
Also, our full algorithm (Fig.\ref{fig:inversion}(e)) shows the best reconstruction quality.
Tab.\ref{tab:reconstruction_quat} gives the quantitative comparison results, where \emph{in-domain} inversion surpasses other competitors from all metrics, including Fréchet Inception Distance (FID) \cite{fid}, Sliced Wasserstein Discrepancy (SWD) \cite{swd}, and Mean-Square Error (MSE).
The inference speed is also shown in Tab.\ref{tab:reconstruction_quat}.
Our \emph{domain-guided} encoder can produce much better reconstruction results compared to the traditional encoder with comparable inference time.
It also provides a better initialization for further the \emph{domain-regularized} optimization, leading to a significantly faster speed ($\sim$35X faster) than the state-of-the-art optimization-based method \cite{image2stylegan}.

\begin{figure}[!ht]
  \centering
  \includegraphics[width=1.0\linewidth]{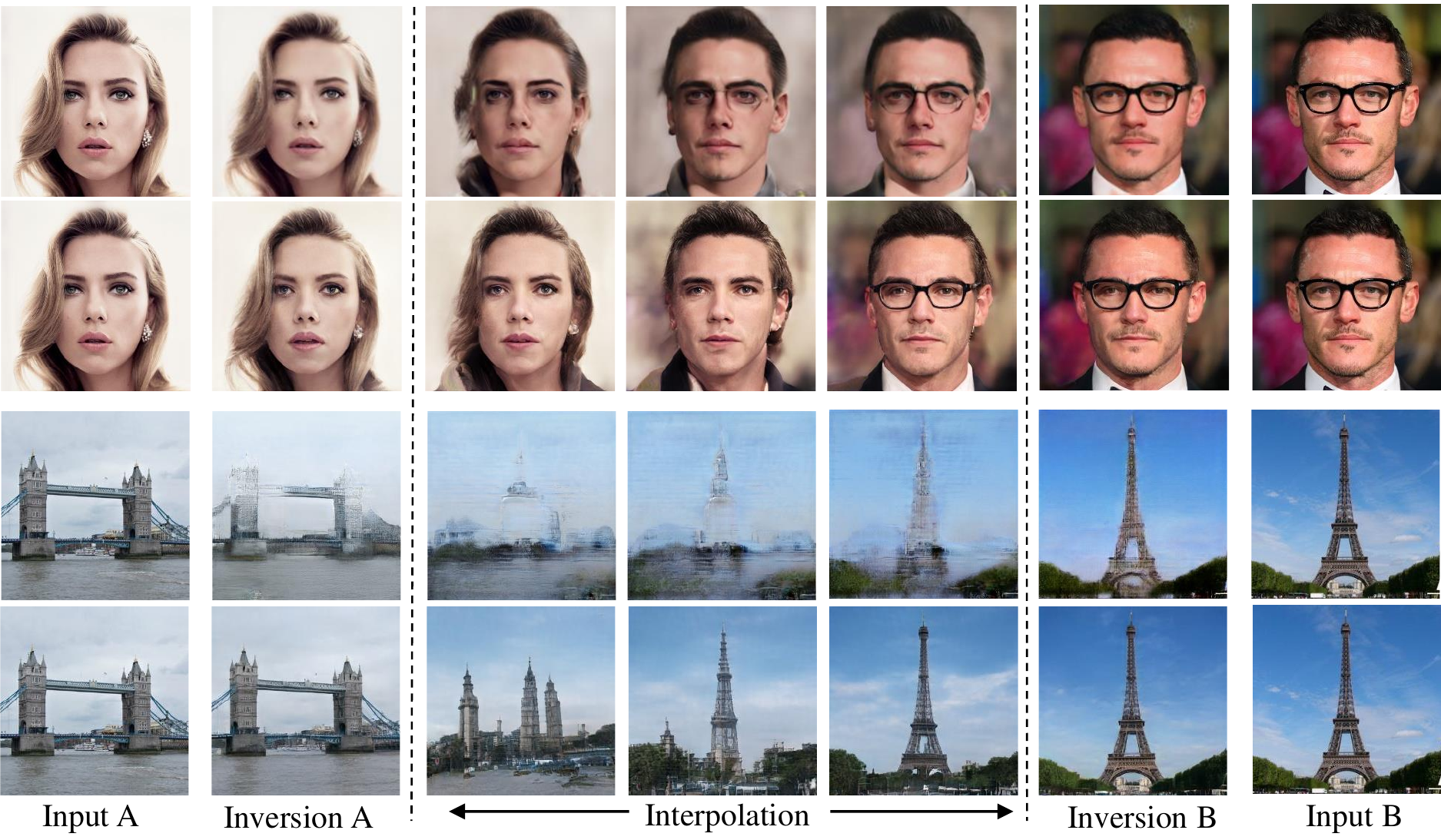}
  \caption{
    Qualitative comparison on image interpolation between Image2StyleGAN \cite{image2stylegan} (odd rows) and our \emph{in-domain} inversion (even rows).
  }
  \label{fig:interpolation}
\end{figure}

\subsection{Real Image Editing}
In this section, we evaluate our \emph{in-domain} GAN inversion approach on real image editing tasks, including image interpolation and semantic image manipulation.
We also come up with a novel image editing task, called \emph{semantic image diffusion}, to see how our approach is able to adapt the content from one image into another and keep the results semantically meaningful and seamlessly compatible.

\noindent\textbf{Image Interpolation.}
Image interpolation aims at semantically interpolating two images, which is suitable for investigating the semantics contained in the inverted codes.
In other words, for a good inversion, the semantic should vary continuously when interpolating two inverted codes.
Fig.\ref{fig:interpolation} shows the comparison results on the image interpolation task between Image2StyleGAN \cite{image2stylegan} and our \emph{in-domain} inversion.
We do experiments on both face and tower datasets to more comprehensively analyze the semantic property.
For the face dataset, our method achieves much smoother interpolated faces than Image2StyleGAN.
For example, in the first two rows of Fig.\ref{fig:interpolation}, eyeglasses are distorted during the interpolation process with Image2StyleGAN and the change from female to male is unnatural.
For tower images, which are much more diverse than faces, the interpolation results from Image2StyleGAN exhibit artifacts and blurriness.
By contrast, our inverted codes lead to more satisfying interpolation.
One noticeable thing is that during interpolating two towers with different types (\emph{e.g.}, one with one spire and the other with multiple spires), the interpolated images using our approach are still high-quality towers.
This demonstrates the \emph{in-domain} property of our algorithm.
Quantitative evaluation in Tab.\ref{tab:editing-comparison} gives the same conclusion.

\setlength{\tabcolsep}{3pt}
\begin{table}[t]
  \caption{
    Quantitative comparison on image interpolation and manipulation between Image2StyleGAN \cite{image2stylegan} and our \emph{in-domain} inversion.
    $\downarrow$ means lower number is better.
  }
  \label{tab:editing-comparison}
  \scriptsize\centering
  \begin{tabular}{|l|cc|cc|cc|cc|}
    \hline
                                                  &                 \multicolumn{4}{c|}{Interpolation}                &                  \multicolumn{4}{c|}{Manipulation}                \\ \hline
                                                  &    \multicolumn{2}{c|}{Face}    &    \multicolumn{2}{c|}{Tower}   &    \multicolumn{2}{c|}{Face}    &    \multicolumn{2}{c|}{Tower}   \\ \hline
                                  \textbf{Method} &           \FID &           \SWD &           \FID &           \SWD &           \FID &           \SWD &           \FID &           \SWD \\ \hline
    MSE-based Optimization \cite{image2stylegan}  &         112.09 &          38.20 &         121.38 &          67.75 &          83.69 &          28.48 &            113 &          52.91 \\
    In-Domain Inversion (Ours)                    & \textbf{91.18} & \textbf{33.91} & \textbf{57.22} & \textbf{28.24} & \textbf{76.43} & \textbf{17.99} & \textbf{57.92} & \textbf{31.50} \\ \hline
  \end{tabular}
\end{table}

\begin{figure}[t]
  \centering
  \includegraphics[width=1.0\linewidth]{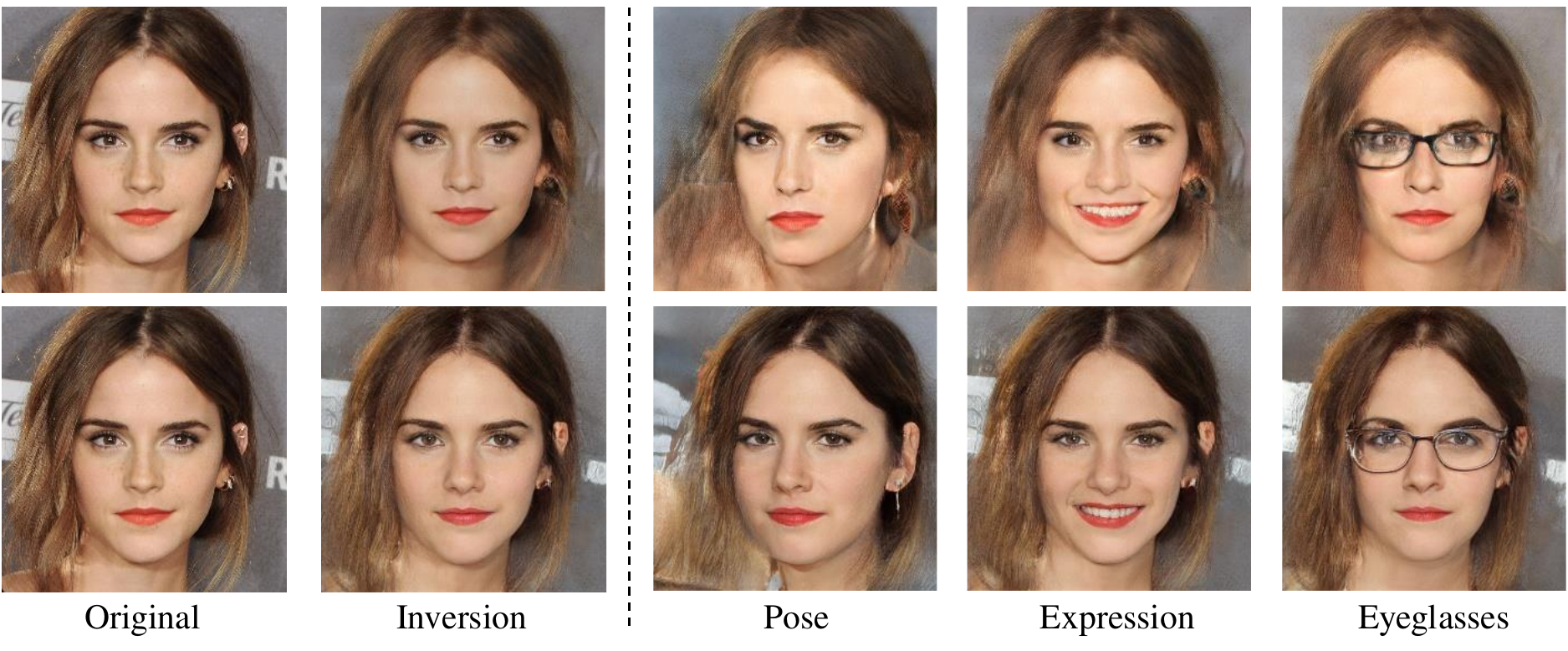}
  \caption{
    Comparison of Image2StyleGAN \cite{image2stylegan} (top row) and our \emph{in-domain} inversion (bottom row) on facial attribute manipulation.
  }
  \label{fig:face-manipulation}
\end{figure}

\begin{figure}[!ht]
  \centering
  \includegraphics[width=1.0\linewidth]{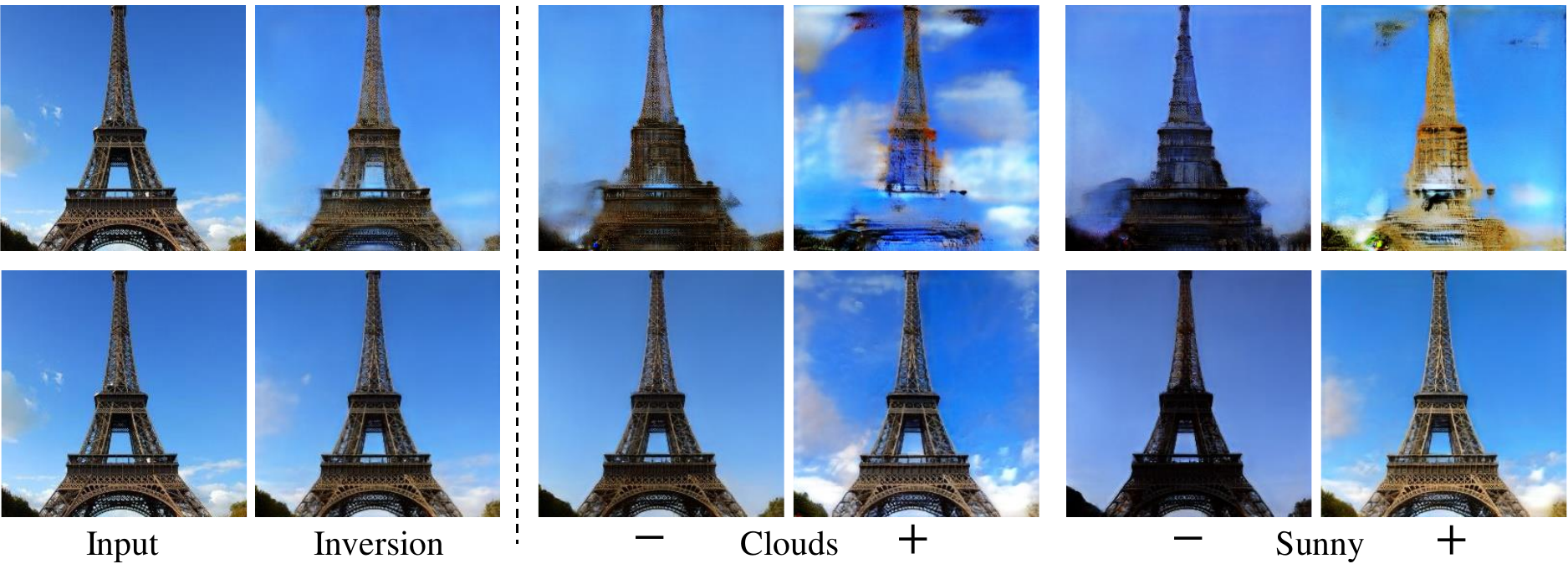}
  \caption{
    Comparison of Image2StyleGAN \cite{image2stylegan} (top row) and our \emph{in-domain} inversion (bottom row) on tower image editing.
    For the manipulation of each attribute, we show the results by either decreasing or increasing the semantic degree.
  }
  \label{fig:tower-manipulation}
\end{figure}

\noindent\textbf{Semantic Manipulation.}
Image manipulation is another way to examine whether the embedded latent codes align with the semantic knowledge learned by GANs.
As pointed out by prior work \cite{shen2019interpreting,yang2019semantic}, GANs can learn rich semantics in the latent space, enabling image manipulation by linearly transforming the latent representation.
This can be formulated as
\begin{align}
  \x^{edit} = G(\z^{inv} + \alpha\n), \label{eq:manipulation}
\end{align}
where $\n$ is the normal direction corresponding to a particular semantic in the latent space and $\alpha$ is the step for manipulation.
In other words, if a latent code is moved towards this direction, the semantics contained in the output image should vary accordingly.
We follow \cite{shen2019interpreting} to search the semantic direction $\n$.

Fig.\ref{fig:face-manipulation} and Fig.\ref{fig:tower-manipulation} show the comparison results of manipulating faces and towers using Images2StyleGAN \cite{image2stylegan} and our \emph{in-domain} GAN inversion.
We can see that our method shows more satisfying manipulation results than Image2StyleGAN.
Taking face manipulation (Fig.\ref{fig:face-manipulation}) as an example, the hair of the actress becomes blurred after the pose rotation using Image2StyleGAN and the identity changes a lot when editing expression and eyeglasses with the codes from Image2StyleGAN.
That is because it only focuses on the reconstruction of the per-pixel values yet omits the semantic information contained in the inverted codes.
On the contrary, our \emph{in-domain} inversion can preserve most other details when editing a particular facial attribute.
As for tower manipulation, we observe from Fig.\ref{fig:tower-manipulation} that our \emph{in-domain} approach surpasses MSE-based optimization by both decreasing and increasing the semantic level.
For example, when removing or adding clouds in the sky, Image2StyleGAN will blur the tower together with the sky, since it only recovers the image at the pixel level without considering the semantic meaning of the recovered objects.
Therefore, the cloud is added to the entire image regardless whether a particular region belongs to sky or tower.
By contrast, our algorithm barely affects the tower itself when editing clouds, suggesting that our \emph{in-domain} inversion can produce semantically informative latent codes for image reconstruction.
We also include the quantitative evaluation on the manipulation task in Tab.\ref{tab:editing-comparison}.
We can tell that our \emph{in-domain} inversion outperforms Image2StyleGAN from all evaluation metrics.

\begin{figure}[t]
  \centering
  \includegraphics[width=1.0\linewidth]{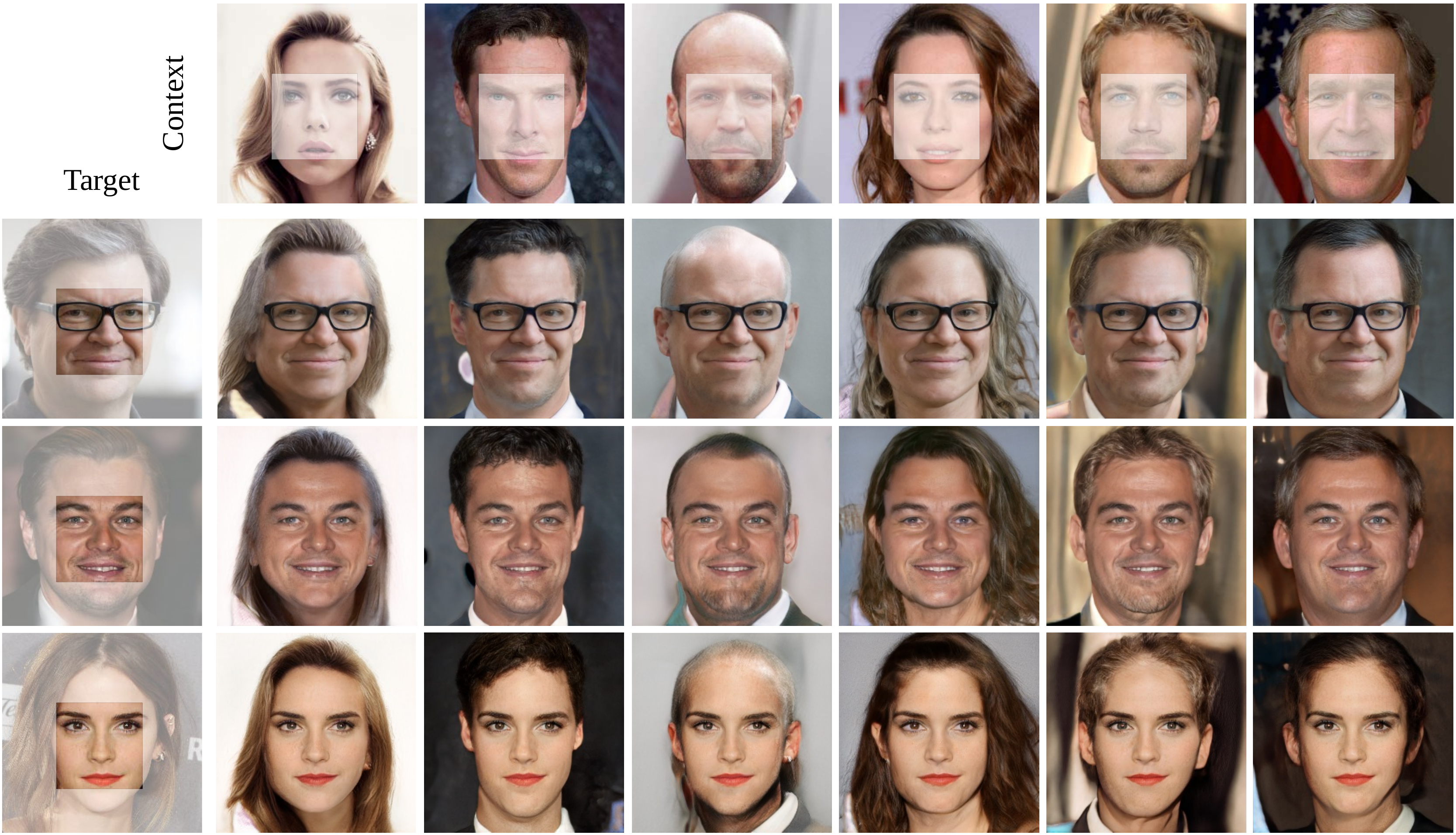}
  \caption{
    Semantic diffusion result using our \emph{in-domain} GAN inversion.
    Target images (first column) are seamlessly diffused into context images (first row) while the identify remains the same as the target.
  }
  \label{fig:face-diffusion}
\end{figure}

\noindent\textbf{Semantic Diffusion.}
Semantic diffusion aims at diffusing a particular part (usually the most representative part) of the target image into the context of another image.
We would like the fused result to keep the characteristics of the target image (\emph{e.g.}, identity of face) and adapt the context information at the same time.
Fig.\ref{fig:face-diffusion} shows some examples where we successfully diffuse various target faces into diverse contexts using our \emph{in-domain} GAN inversion approach.
We can see that the results well preserve the identity of the target face and reasonably integrate into different surroundings.
This is different from style mixing since the center region of our resulting image is kept the same as that of the target image.
More detailed analysis on the semantic diffusion operation can be found in the \textbf{Appendix}.

\begin{figure}[t]
  \centering
  \includegraphics[width=1.0\linewidth]{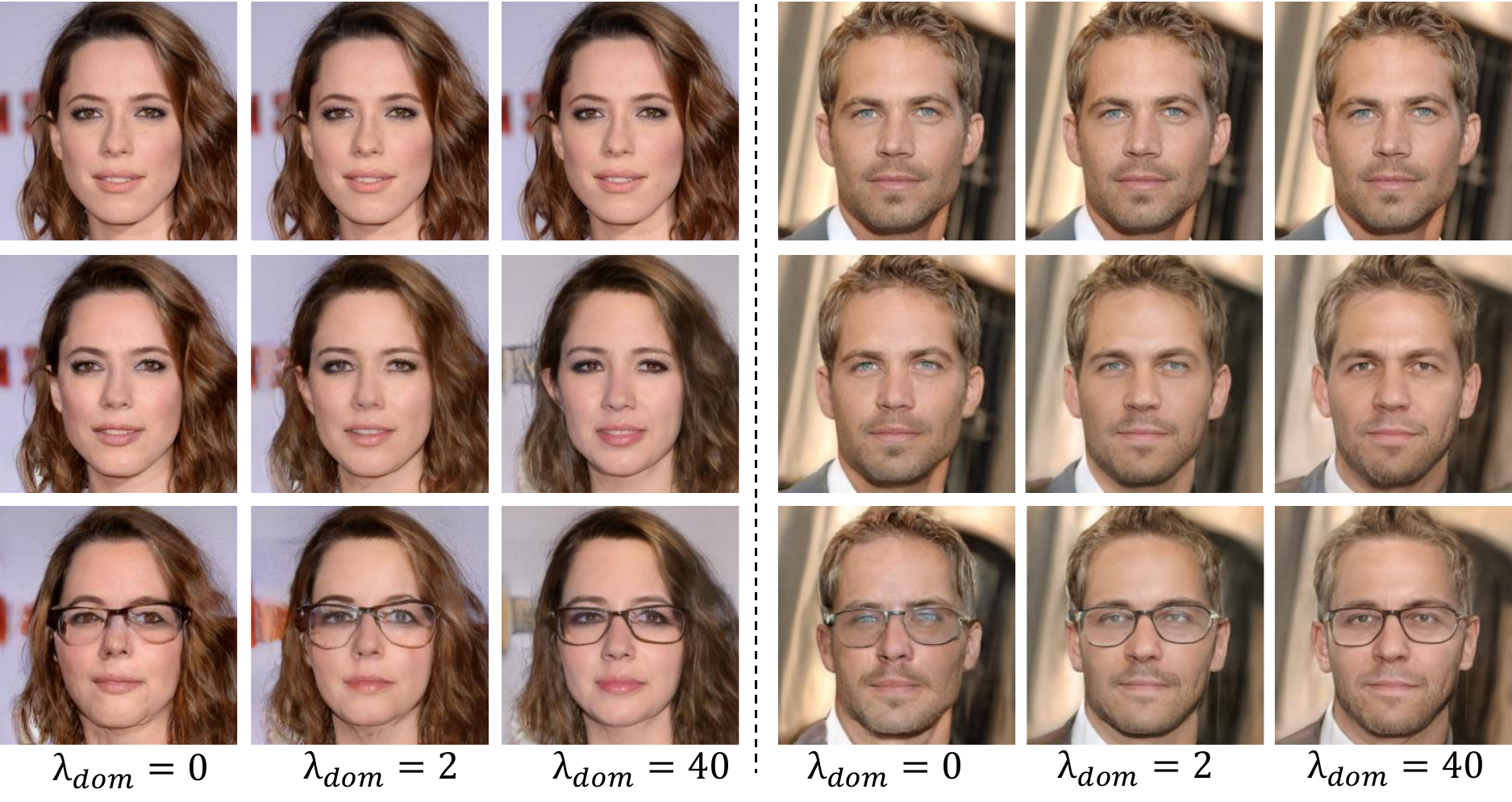}
  \caption{
    Ablation study on the loss weight in Eq.\eqref{eq:optimization} for the \emph{domain-regularized} optimization.
    From top to bottom: original images, reconstructed images, and manipulation results (wearing eyeglasses).
    For each group of images, the weight $\lambda_{dom}$ is set to be 0, 2, 40.
    When $\lambda_{dom}$ equals to 0, it produces the best reconstructed results but relatively poor manipulation results.
    When $\lambda_{dom}$ equals to 40, we get worse reconstruction but more satisfying manipulation.
  }
  \label{fig:encoder-reg-weight}
\end{figure}

\begin{figure}[t]
  \centering
  \includegraphics[width=0.75\linewidth]{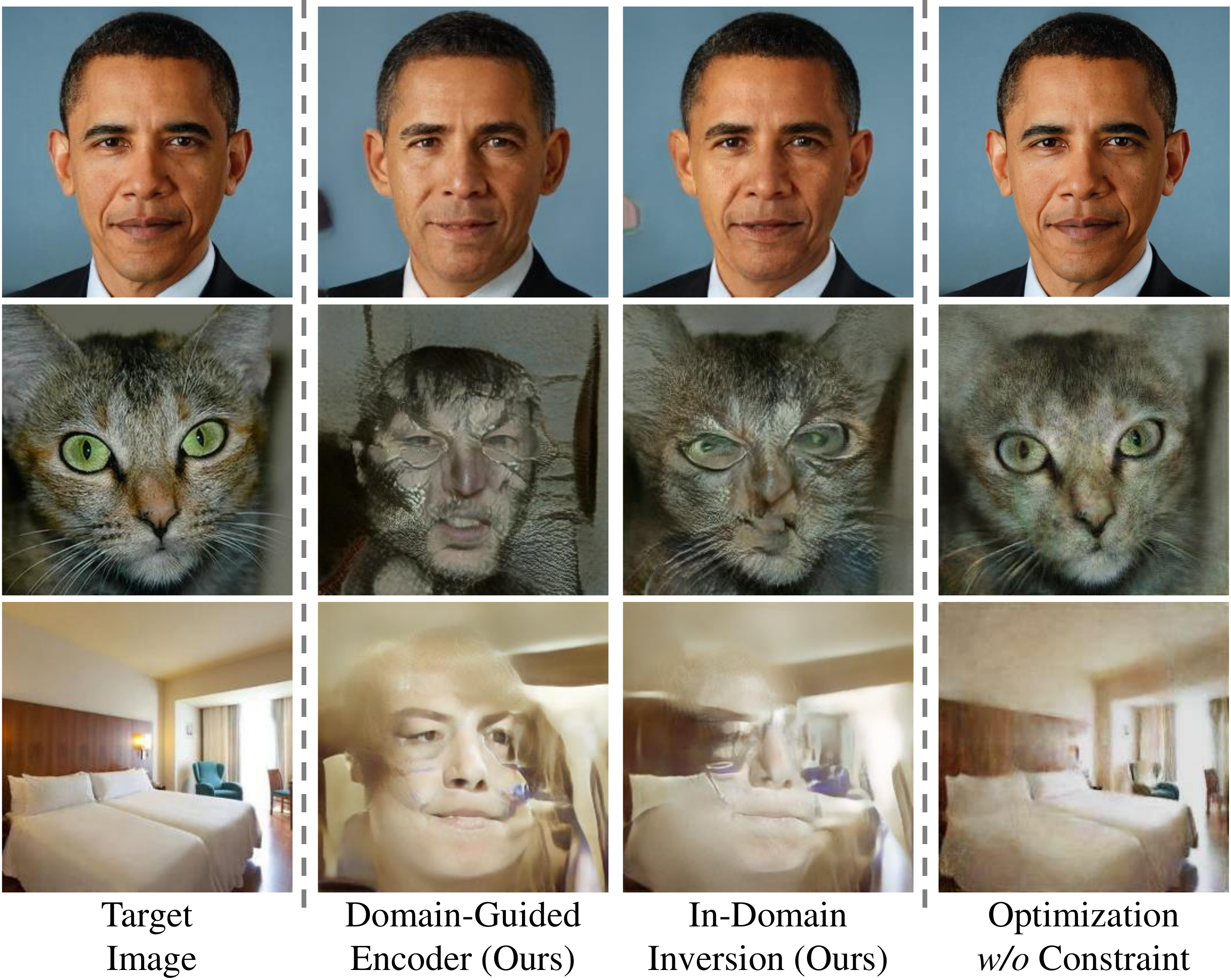}
  \caption{
    Results on inverting face, cat face, and bedroom using the same face synthesis model.
    From left to right: target images, reconstruction results with the outputs from the \emph{domain-guided} encoder, reconstruction results with the proposed \emph{in-domain} inversion, reconstruction results by directly optimizing the latent code \emph{w/o} considering domain alignment \cite{image2stylegan}.
  }
  \label{fig:cross-domain}
\end{figure}

\subsection{Ablation Study}
In this part, we conduct an ablation study to analyze the proposed \emph{in-domain} inversion.
After the initial training of the encoder, we perform the \emph{domain-regularized} optimization on each image to further improve the reconstruction quality.
Different from the previous MSE-based optimization, we involve the learned \emph{domain-guided} encoder as a regularizer to land the inverted code inside the semantic domain, as described in Eq.\eqref{eq:optimization}.
Here, we study the role of the encoder in the optimization process by varying the weight $\lambda_{dom}$ in Eq.\eqref{eq:optimization}.
Fig.\ref{fig:encoder-reg-weight} shows the comparison between $\lambda_{dom} = 0, 2, 40$.
We observe the trade-off between the image reconstruction quality and the manipulation quality.
Larger $\lambda_{dom}$ will bias the optimization towards the domain constraint such that the inverted codes are more semantically meaningful.
Instead, the cost is that the target image cannot be ideally recovered for per-pixel values.
In practice, we set $\lambda_{dom} = 2$.

\section{Discussion and Conclusion}\label{sec:conclusion}
In this work, we explore the \emph{semantic} property of the inverted codes in the GAN inversion task and propose a novel \emph{in-domain} inversion method.
To the best of our knowledge, this is the first attempt to invert a pre-trained GAN model \emph{explicitly} considering the semantic knowledge encoded in the latent space.
We show that the code that simply recovers the pixel value of the target image is not sufficient to represent the image at the semantic level.
For example, in Fig.\ref{fig:cross-domain}, we invert different types of image instances (\emph{i.e.}, face, cat face, and bedroom) with the face synthesis model.
The last column shows the results from Image2StyleGAN \cite{image2stylegan} which recovers a cat or a bedroom with the domain knowledge learned to synthesis human faces.
By contrast, the face outline can still be observed in the reconstructions using our \emph{in-domain} inversion (third column).
This demonstrates, from a different angle, the superiority of our approach in producing semantically meaningful codes.
Taking inverting bedroom (third row) as an example, the bedroom image is outside the domain of the training data and the GAN model should not be able to learn the bedroom-related semantics.
Accordingly, reusing the face knowledge to represent a bedroom is ill-defined.
Even though we can always use more parameters to over-fit the pixel values of the bedroom (\emph{e.g.}, the last column), such over-fitting would fail to support semantic image manipulation.
From this viewpoint, our \emph{in-domain} inversion lands the inverted code inside the original domain to make it semantically meaningful.
In other words, we aim at finding the most adequate code to recover the target image from \emph{both the pixel level and the semantic level}.
Such \emph{in-domain} inversion significantly facilitates real image editing.

\noindent\textbf{Acknowledgement.}
This work is supported in part by the Early Career Scheme (ECS) through the Research Grants Council (RGC) of Hong Kong under Grant No.24206219, CUHK FoE RSFS Grant, and SenseTime Collaborative Grant.

\bibliographystyle{splncs04}
\bibliography{ref}

\clearpage
\appendix
\section*{Appendix}

\section{Overview}\label{appendix:overview}
This appendix is organized as follows:
In Sec.\ref{appendix:image-reconstruction}, we show image reconstruction results using the model trained on LSUN bedroom dataset \cite{lsun}.
In Sec.\ref{appendix:image-interpolation} and Sec.\ref{appendix:semantic-manipulation}, we show more results of image interpolation and image manipulation to verify that \emph{in-domain} GAN inversion can recover the target images from both the pixel level and the semantic level.
In Sec.\ref{appendix:style-mixing}, we show some style mixing results.
In Sec.\ref{appendix:diffusion}, we make detailed analysis on the semantic diffusion achieved by our \emph{in-domain} inversion.

\section{Image Reconstruction}\label{appendix:image-reconstruction}
\begin{figure}[ht]
  \centering
  \includegraphics[width=0.98\linewidth]{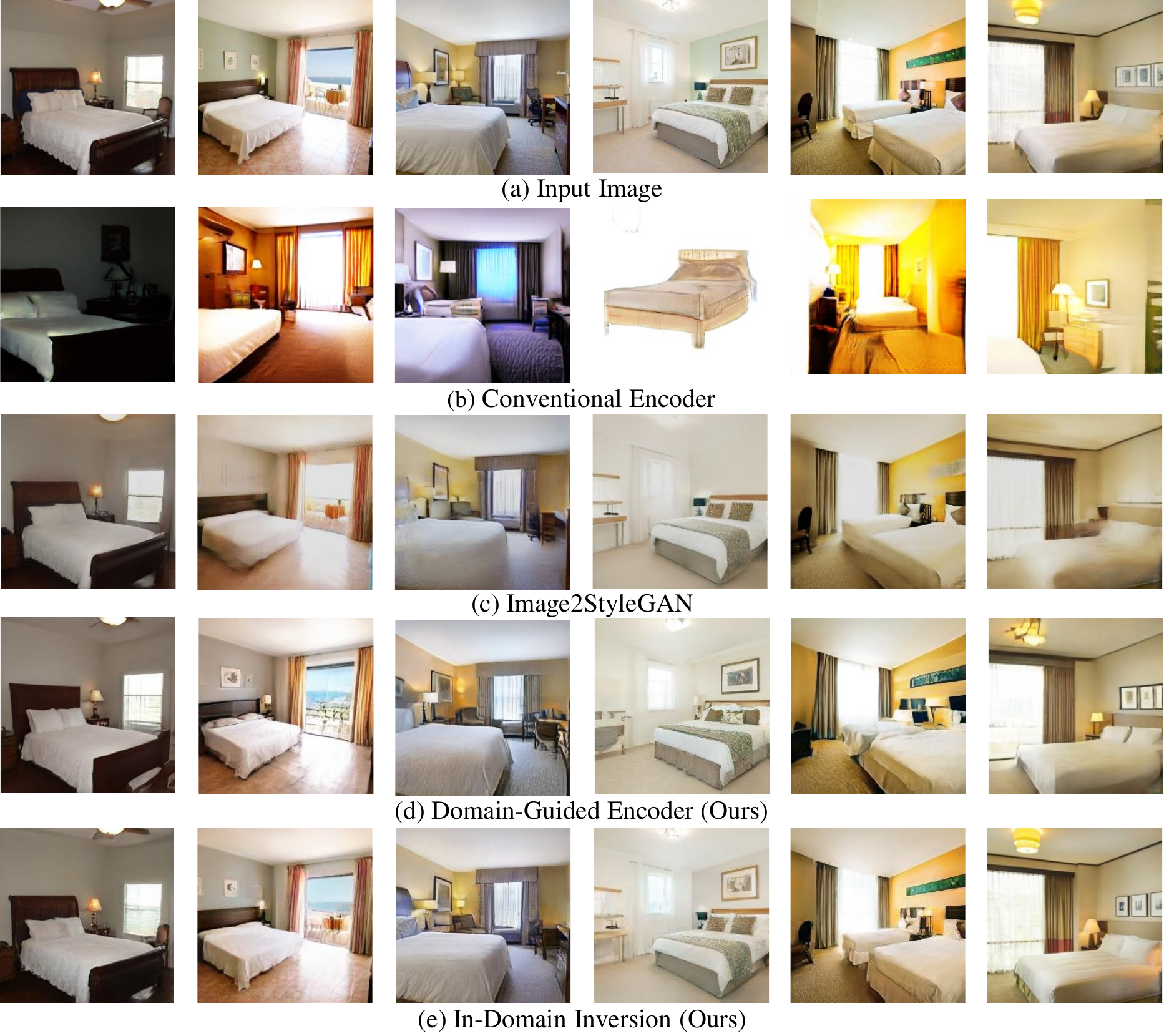}
  \caption{
    Qualitative comparison on bedroom reconstruction with different GAN inversion methods.
    (a) Input image.
    (b) Conventional encoder \cite{zhu2016generative}.
    (c) Image2StyleGAN \cite{image2stylegan}.
    (d) Our proposed \emph{domain-guided} encoder.
    (e) Our proposed \emph{in-domain} inversion.
  }
  \label{appendix:fig:inversion}
\end{figure}

Image reconstruction is one of the most important metrics to evaluate a GAN inversion method.
Besides human faces and towers (outdoor scene) shown in the main paper, we also do experiments on bedrooms (indoor scene) and compare with existing inversion methods.
The comparison results are shown in Fig.\ref{appendix:fig:inversion}.
We can tell that our proposed \emph{domain-guided} encoder produces much better reconstructions than the conventional encoder \cite{zhu2016generative}.
The further \emph{domain-regularized} optimization also surpasses the start-of-the-art optimization-based inversion method, Image2StyleGAN \cite{image2stylegan}, with higher reconstruction quality.

\section{Image Interpolation}\label{appendix:image-interpolation}
Different from previous GAN inversion approaches that mainly focus on the image reconstruction from the pixel level, we propose to align the inverted code with the semantic knowledge learn by GAN models, \emph{i.e.}, \emph{in-domain}.

\begin{figure}[t]
  \centering
  \includegraphics[width=1.0\linewidth]{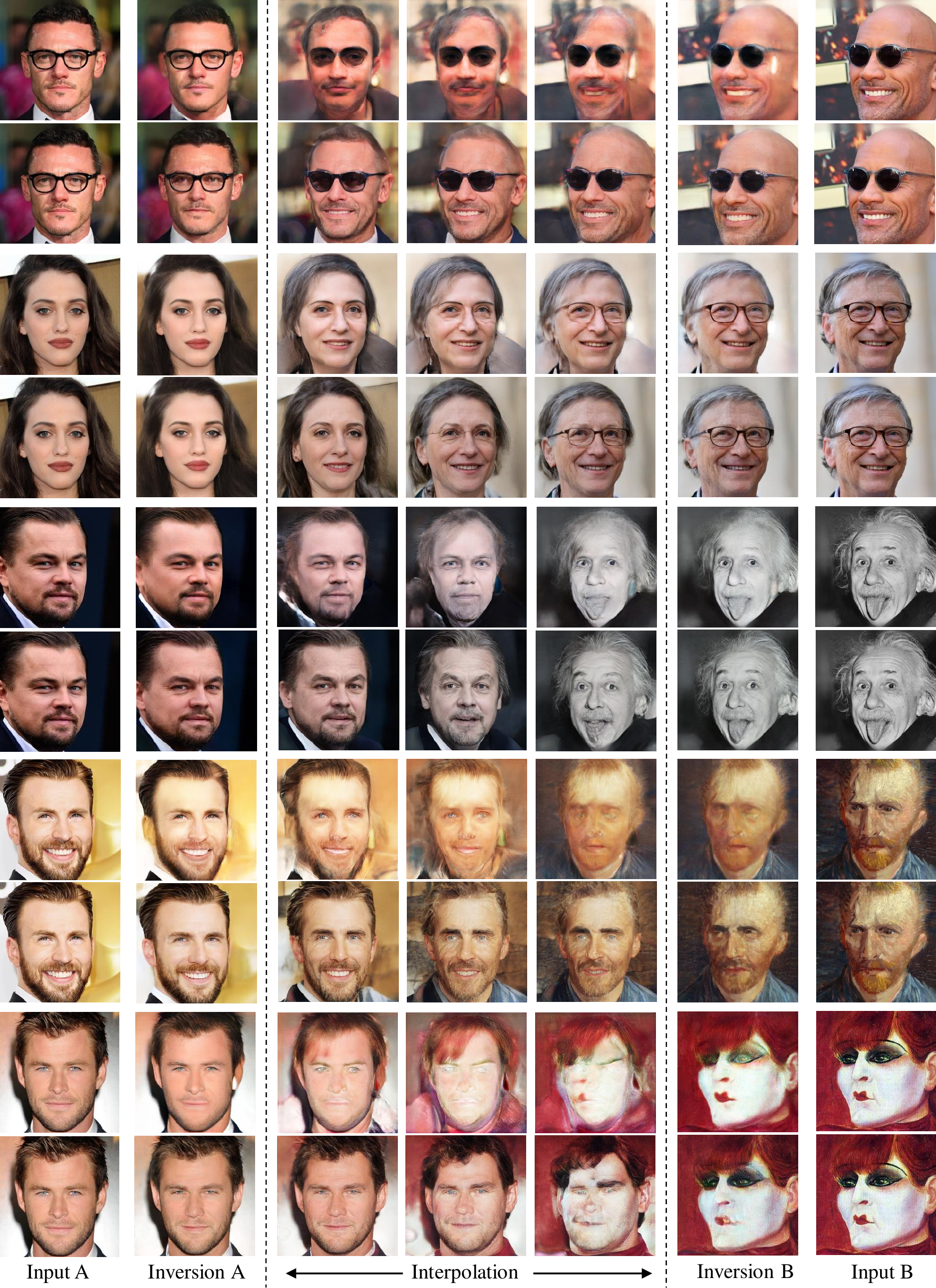}
  \caption{
    Qualitative comparison on face interpolation between Image2StyleGAN \cite{image2stylegan} (odd rows) and our \emph{in-domain} inversion (even rows).
    Zoom in for details.
  }
  \label{appendix:fig:interpolation-face-compare}
\end{figure}

\begin{figure}[t]
  \centering
  \includegraphics[width=1.0\linewidth]{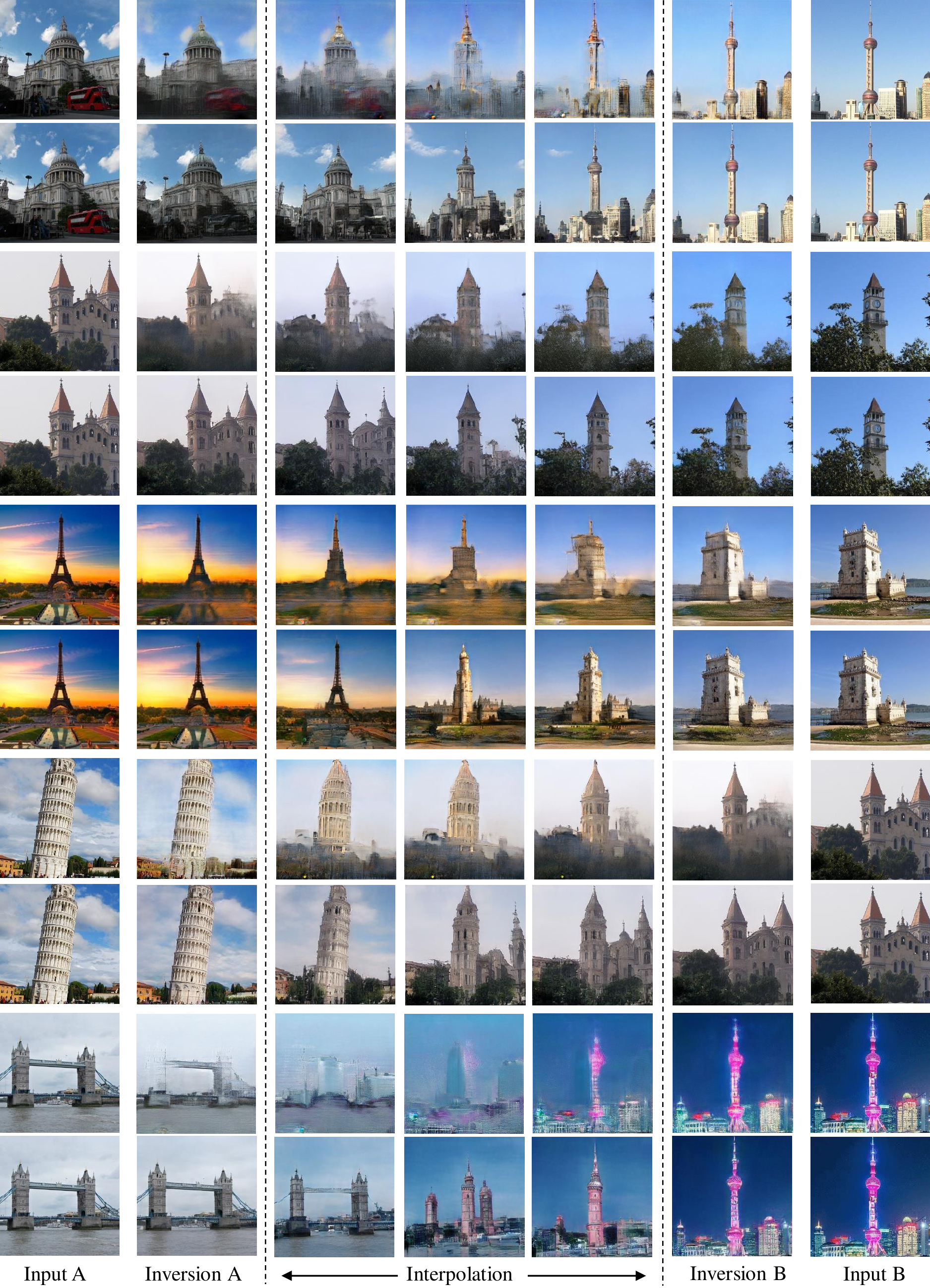}
  \caption{
    Qualitative comparison on tower interpolation between Image2StyleGAN \cite{image2stylegan} (odd rows) and our \emph{in-domain} inversion (even rows).
    Zoom in for details.
  }
  \label{appendix:fig:interpolation-tower-compare}
\end{figure}

\begin{figure}[t]
  \centering
  \includegraphics[width=1.0\linewidth]{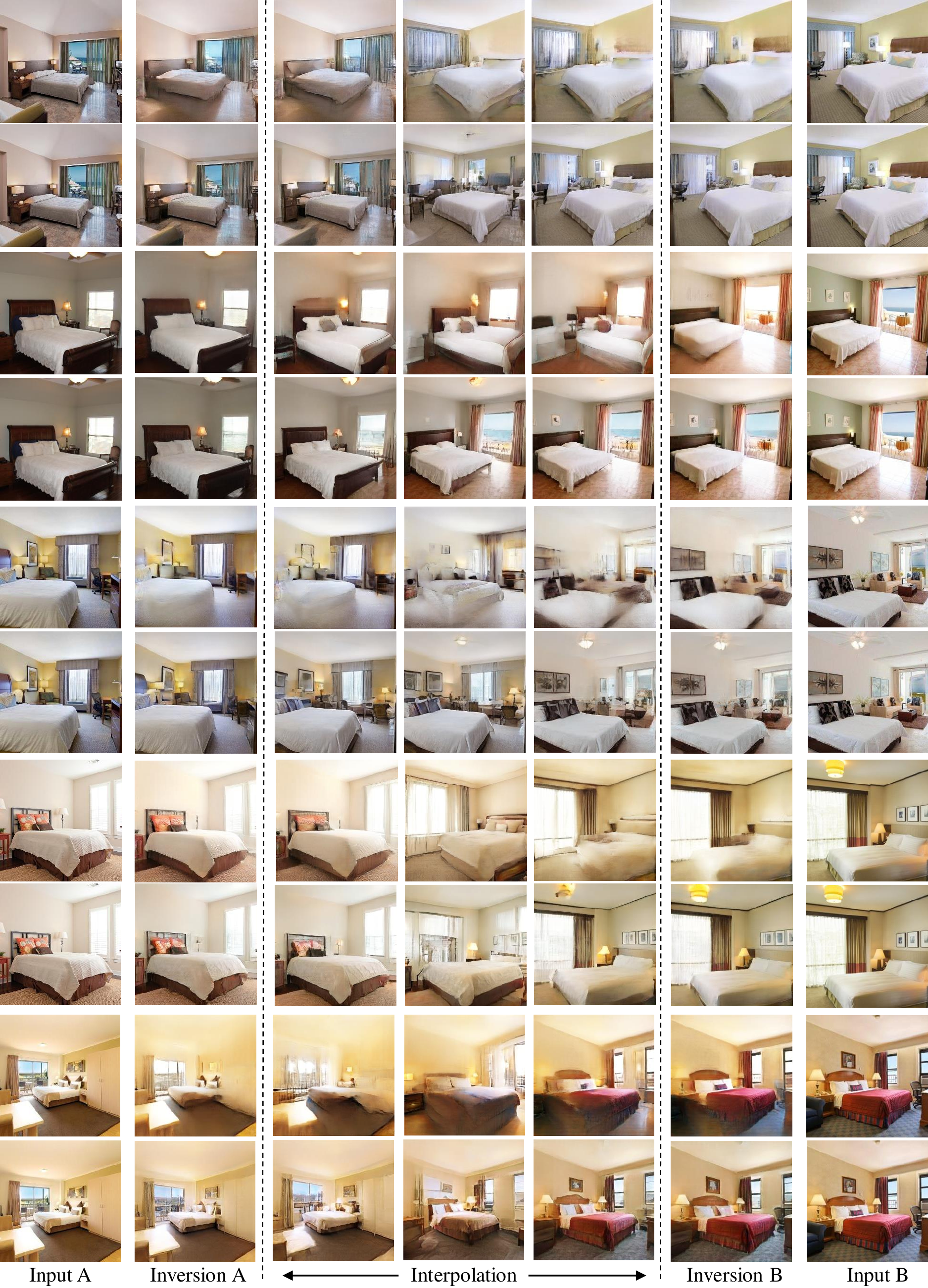}
  \caption{
     Qualitative comparison on bedroom interpolation between Image2StyleGAN \cite{image2stylegan} (odd rows) and our \emph{in-domain} inversion (even rows).
    Zoom in for details.
  }
  \label{appendix:fig:interpolation-bedroom-compare}
\end{figure}

\begin{figure}[t]
  \centering
  \includegraphics[width=1.0\linewidth]{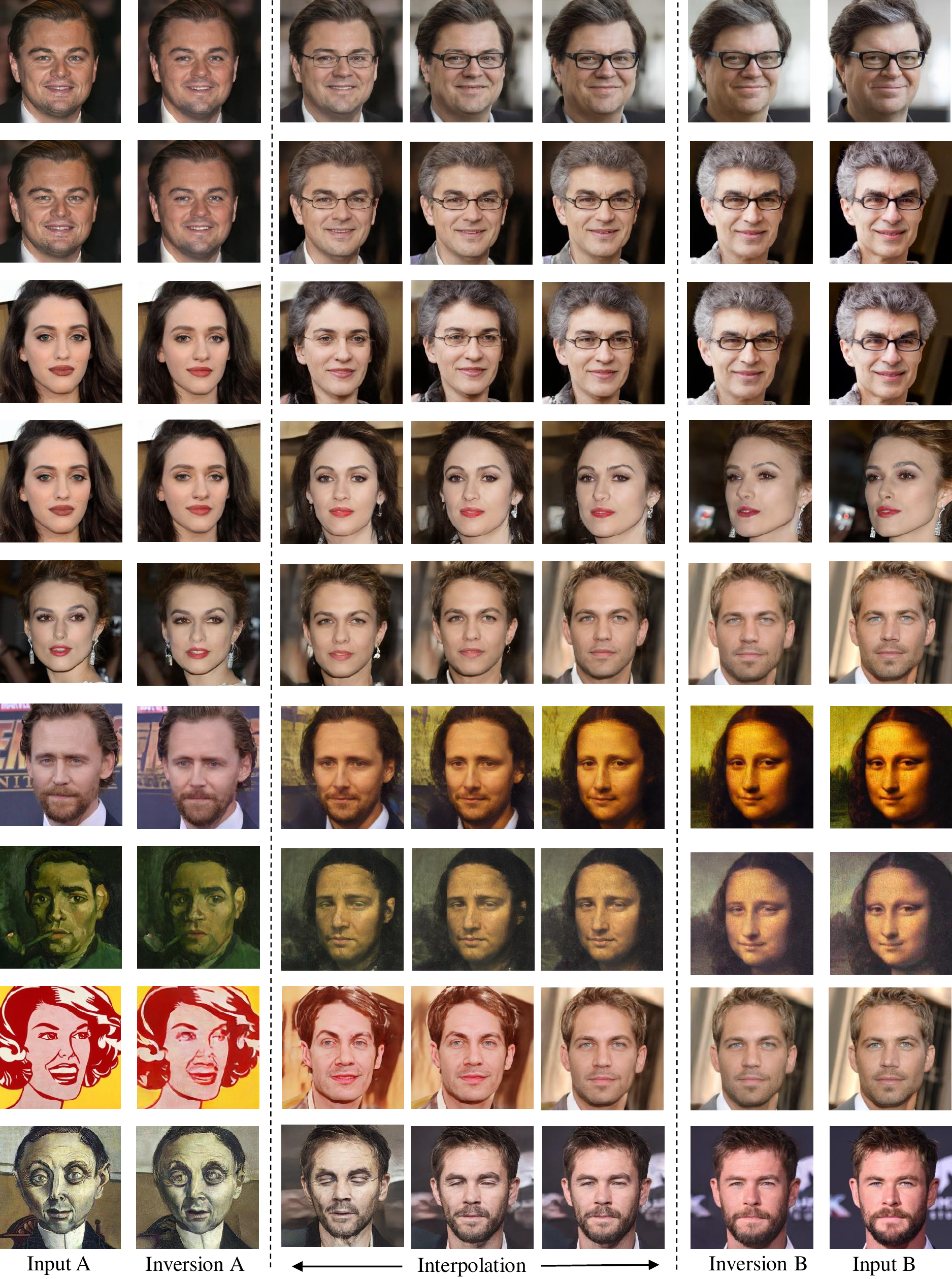}
  \caption{
    Face interpolation results using our \emph{in-domain} GAN inversion method.
    Zoom in for details.
  }
  \label{appendix:fig:interpolation-face}
\end{figure}

\begin{figure}[t]
  \centering
  \includegraphics[width=1.0\linewidth]{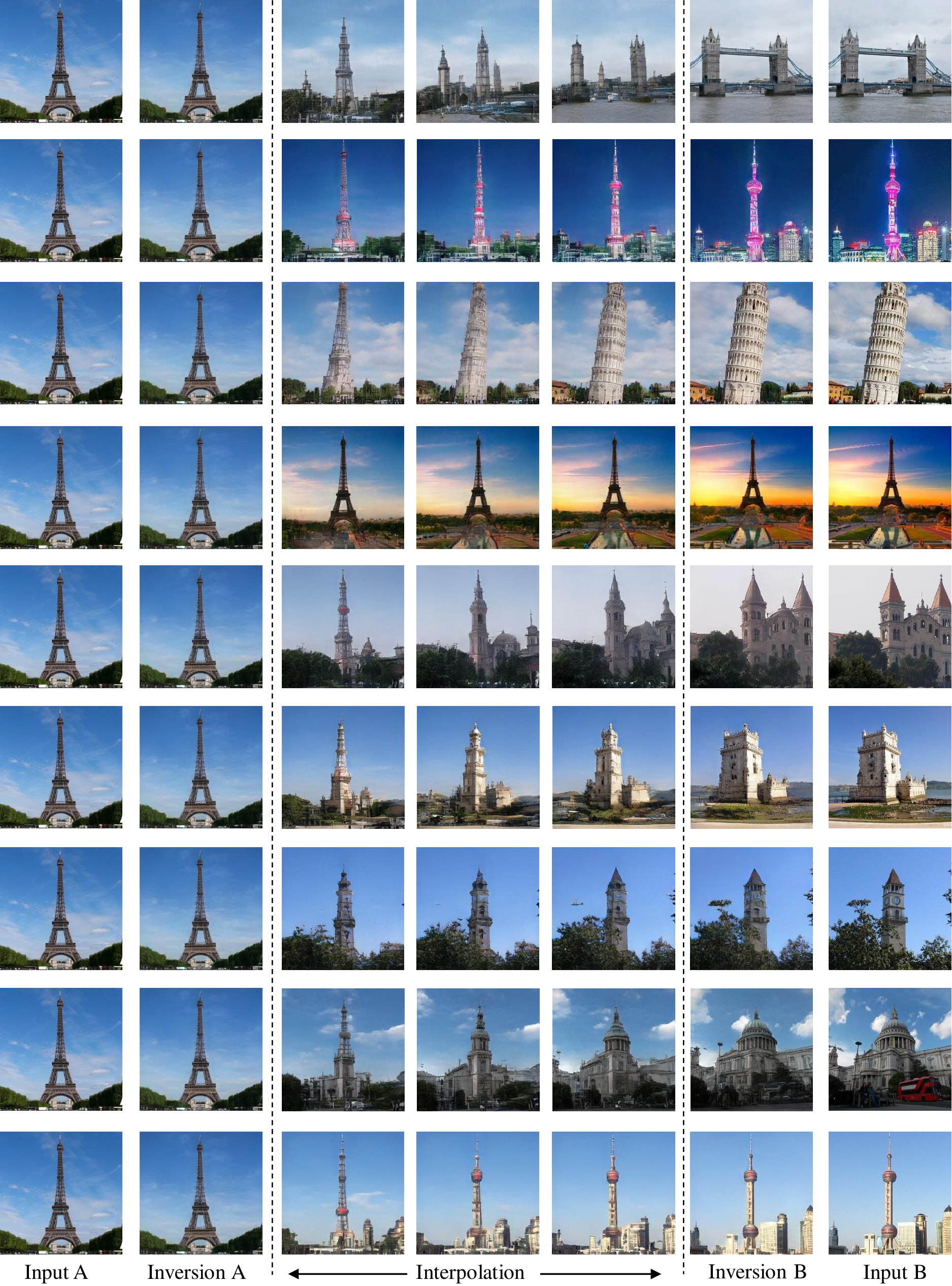}
  \caption{
    Tower interpolation results using our \emph{in-domain} GAN inversion method.
    Zoom in for details.
  }
  \label{appendix:fig:interpolation-tower}
\end{figure}

\begin{figure}[t]
  \centering
  \includegraphics[width=1.0\linewidth]{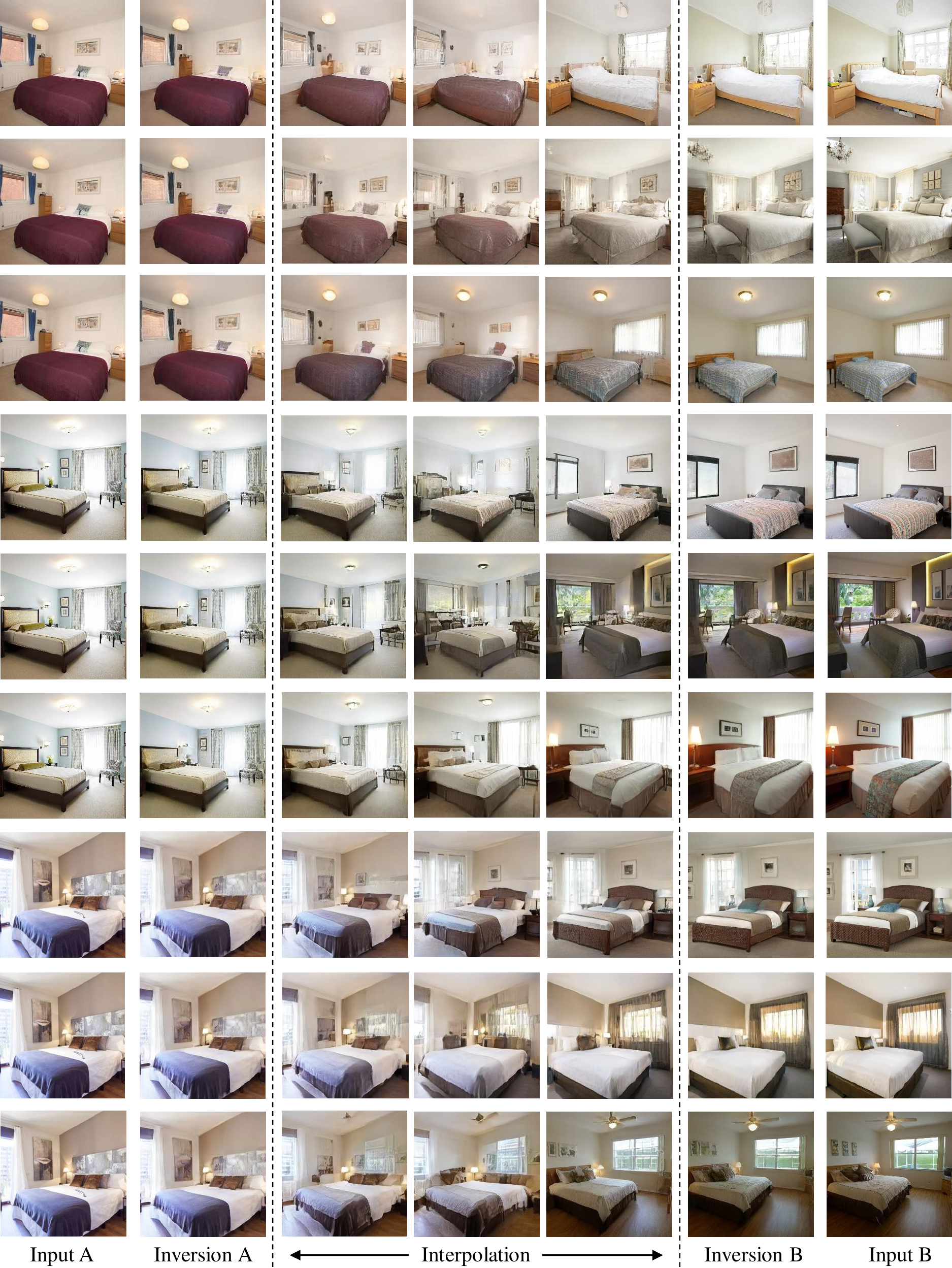}
  \caption{
    Bedroom interpolation results using our \emph{in-domain} GAN inversion method.
    Zoom in for details.
  }
  \label{appendix:fig:interpolation-bedroom}
\end{figure}

In this section, we use image interpolation to evaluate whether the inverted codes are semantically meaningful.
Fig.\ref{appendix:fig:interpolation-face-compare}, Fig.\ref{appendix:fig:interpolation-tower-compare}, and Fig.\ref{appendix:fig:interpolation-bedroom-compare} show the comparison results between Image2StyleGAN \cite{image2stylegan} and our \emph{in-domain} inversion on faces, towers (outdoor scene), and bedrooms (indoor scene) respectively.
We observe that the interpolations from Image2StyleGAN show unsatisfying artifacts and blurs, especially when the source and target images are with large discrepancy (\emph{e.g.}, the first and last sample in Fig.\ref{appendix:fig:interpolation-face-compare}).
Meanwhile, some interpolations made by Image2StyleGAN are not semantically meaningful (\emph{e.g.}, interpolated images are no longer a tower any more in the last sample in Fig.\ref{appendix:fig:interpolation-tower-compare}).
%
%
On the contrary, our method makes sure that all interpolated samples are still with high quality and explanatory semantics.
We show more results in Fig.\ref{appendix:fig:interpolation-face} (face), Fig.\ref{appendix:fig:interpolation-tower} (tower), and  Fig.\ref{appendix:fig:interpolation-bedroom} (bedroom).
In Fig.\ref{appendix:fig:interpolation-face}, we manage to interpolate male and female, faces with different poses, or even painting and real person.
In Fig.\ref{appendix:fig:interpolation-tower}, we interpolate one type of tower to other types in a large diversity.
Each individual interpolation is realistic enough for a ``new type'' of tower.
In Fig.\ref{appendix:fig:interpolation-bedroom}, we can interpolate between bedrooms from different viewpoints.
It is also noteworthy that windows and paintings on the wall can also be adequately interpolated using our method.

\clearpage

\section{Semantic Manipulation}\label{appendix:semantic-manipulation}
\begin{figure}[t]
  \centering
  \includegraphics[width=1.0\linewidth]{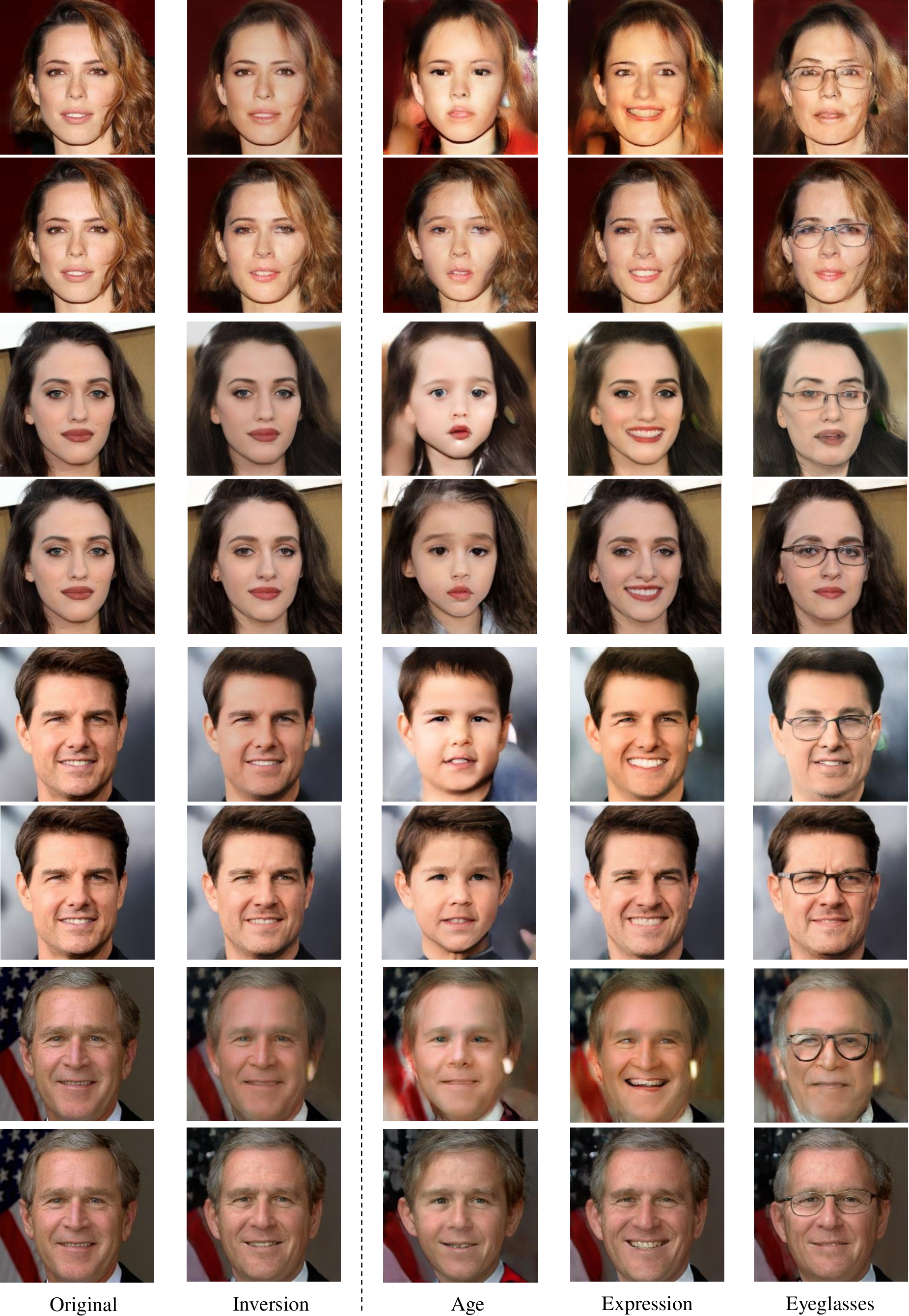}
  \caption{
    Comparison results on manipulating face images between Image2StyleGAN \cite{image2stylegan} and our \emph{in-domain} GAN inversion.
  }
  \label{appendix:fig:manipulation-face}
\end{figure}

\begin{figure}[t]
  \centering
  \includegraphics[width=1.0\linewidth]{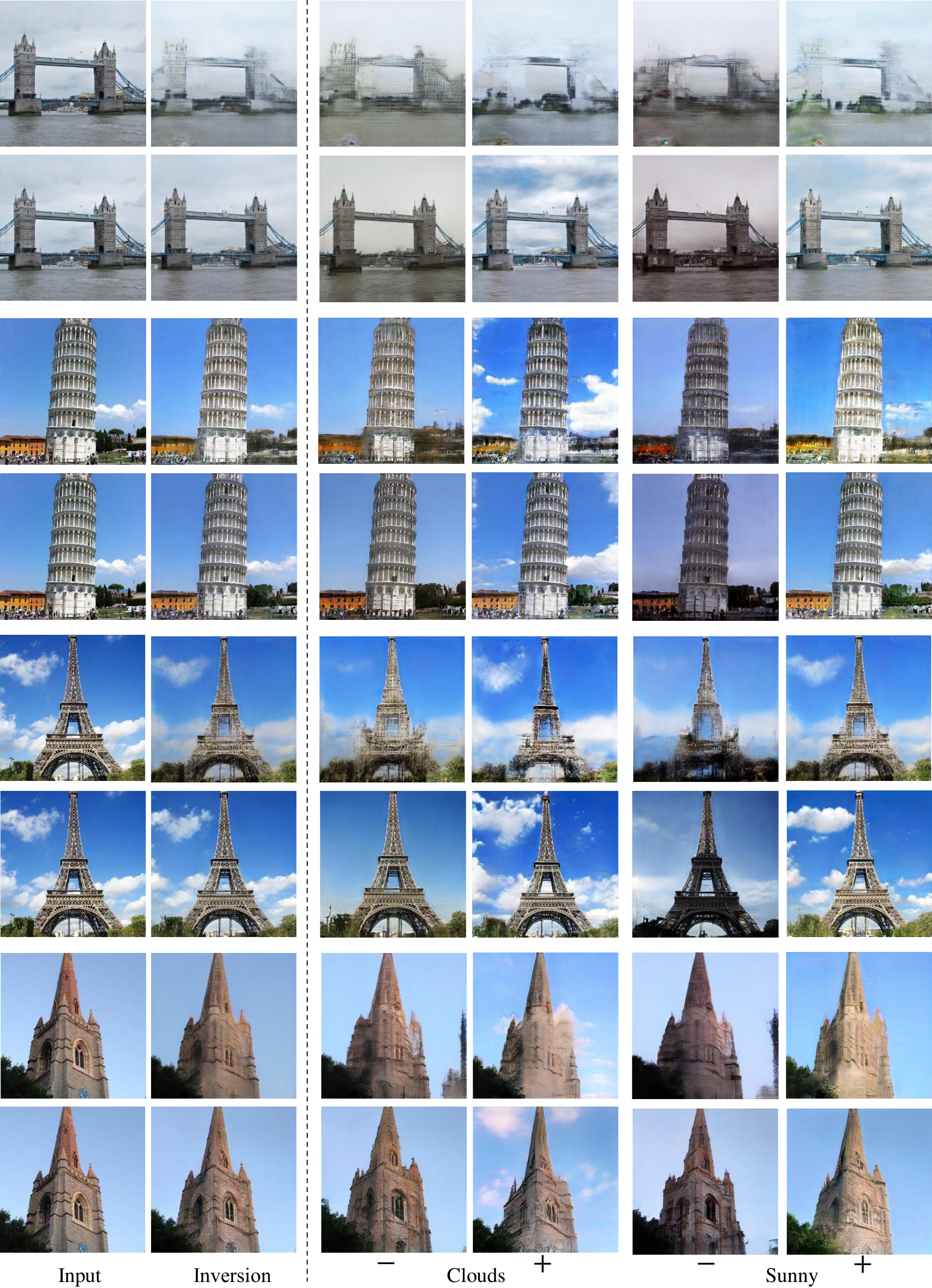}
  \caption{
    Comparison results on manipulating tower images between Image2StyleGAN \cite{image2stylegan} and our \emph{in-domain} GAN inversion.
  }
  \label{appendix:fig:manipulation-tower}
\end{figure}

\begin{figure}[t]
  \centering
  \includegraphics[width=1.0\linewidth]{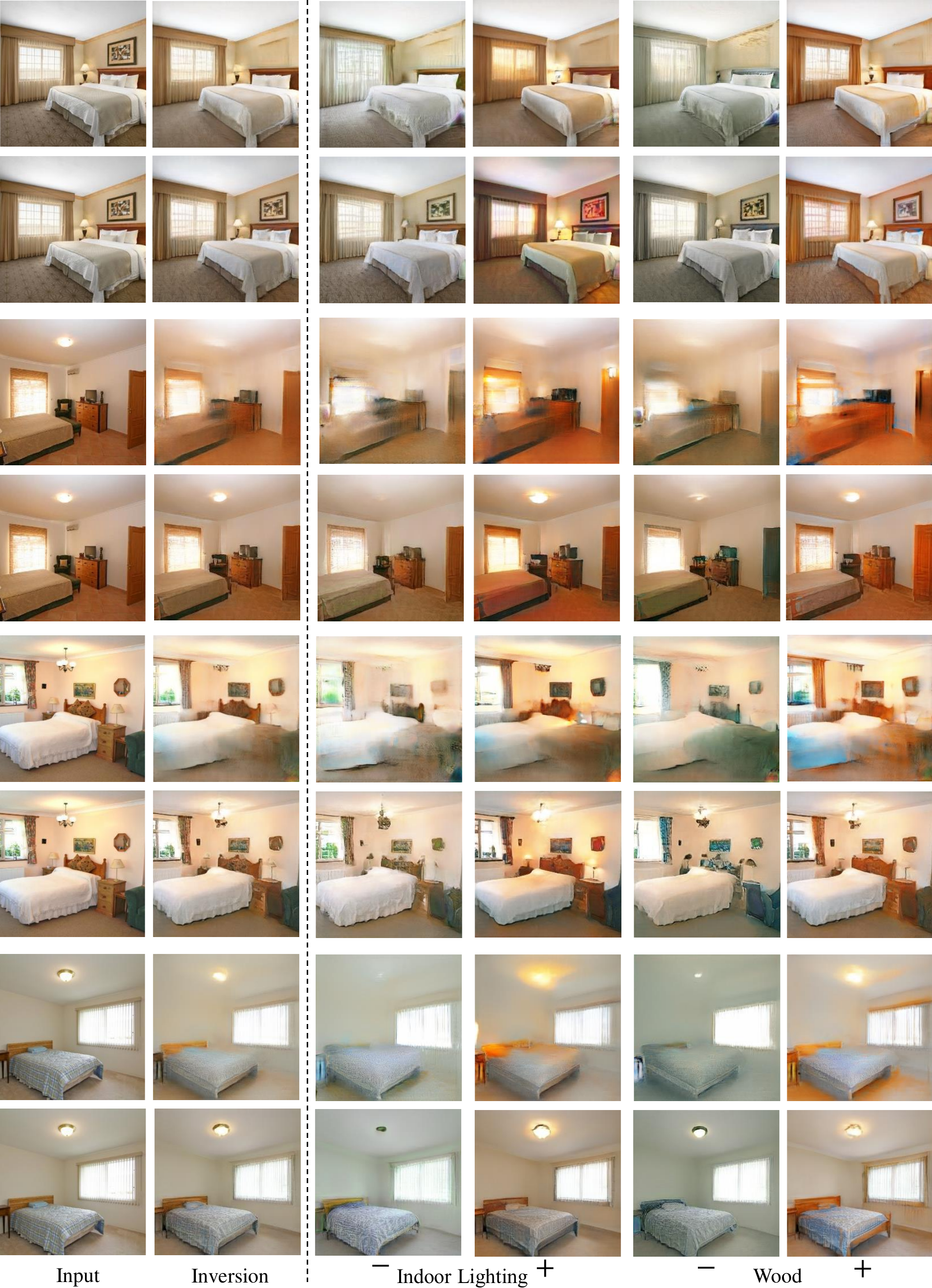}
  \caption{
    Comparison results on manipulating bedroom images between Image2StyleGAN \cite{image2stylegan} and our \emph{in-domain} GAN inversion.
  }
  \label{appendix:fig:manipulation-bedroom}
\end{figure}

Prior work has shown that a well-trained GAN model is able to encode interpretable semantics inside the latent space \cite{shen2019interpreting,gansteerability,yang2019semantic}.
These learned semantics can be further used for real image manipulation together with GAN inversion.
In this section, we compare our \emph{in-domain} inversion with Image2StyleGAN \cite{image2stylegan} on the semantic manipulation task.
Results are shown in Fig.\ref{appendix:fig:manipulation-face} (face), Fig.\ref{appendix:fig:manipulation-tower} (tower), and Fig.\ref{appendix:fig:manipulation-bedroom} (bedroom).
It turns out that we can achieve impressive semantic editing with respect to various attributes, significantly surpassing Image2StyleGAN which usually produces results with artifacts.
That is because the code inverted by Image2StyleGAN is not aligned with the rich semantics encoded in the latent space.
In other words, only trying to recover the pixel values does not support semantically meaningful image editing.
On the contrary, our proposed \emph{in-domain} inversion is able to better reuse the semantic knowledge learned by GANs.

\clearpage

\section{Style Mixing}\label{appendix:style-mixing}
\begin{figure}[t]
  \centering
  \includegraphics[width=1.0\linewidth]{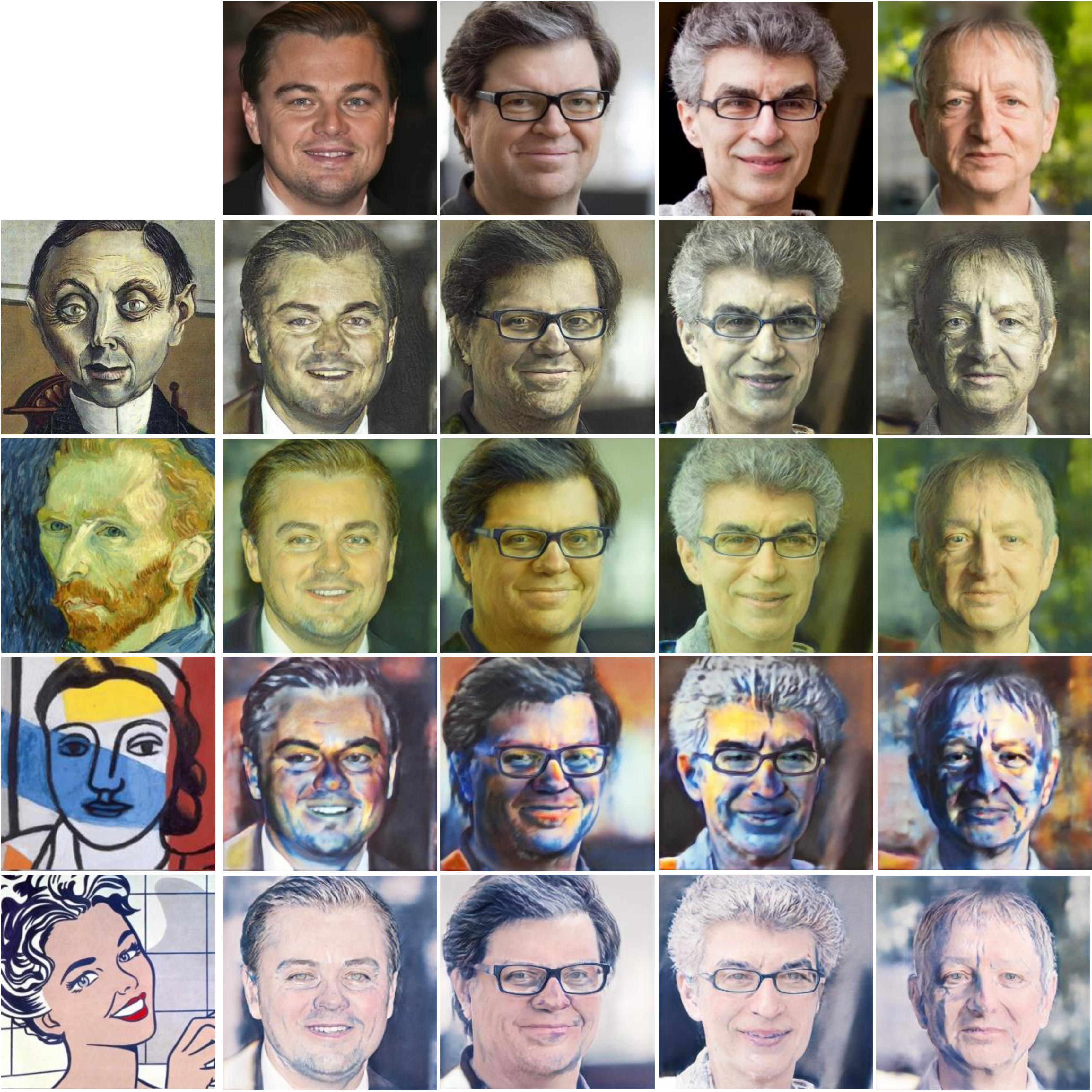}
  \caption{
    Style mixing results using our \emph{in-domain} GAN inversion method.
    First column indicates style images and first row shows content images.
  }
  \label{appendix:fig:style-mixing}
\end{figure}

We also evaluate our approach on the style mixing task, which aims at transferring the style of a style image to a content image.
For this purpose, we invert both the style image and the content image to layer-wise latent codes.
Then, we replace the codes from the last four layers of the content image with those from the style image.
Fig.\ref{appendix:fig:style-mixing} shows the mixing results.
We can tell that each mixture successfully inherits painting style from the artistic face (first column) yet maintains most details from the real person (first row).
This suggests that our \emph{in-domain} inversion manages to convert input images to semantically meaningful latent codes.

\clearpage

\begin{figure}[t]
  \centering
  \includegraphics[width=1.0\linewidth]{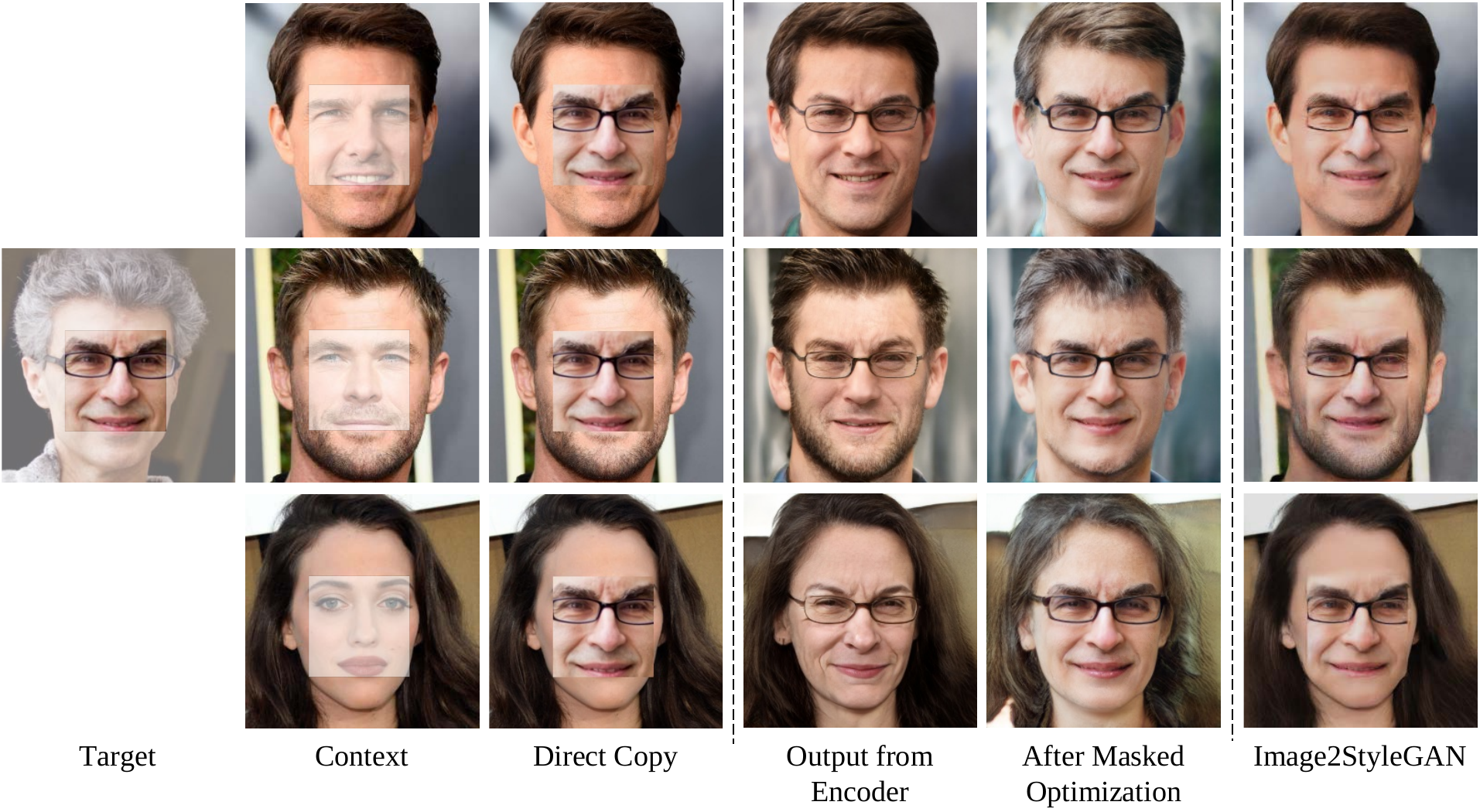}
  \caption{
    The reconstruction from the output of our \emph{domain-guided} encoder already has a smooth transition between the target and context.
    After the masked optimization, the identity of the target is further preserved.
    As a comparison, Image2StyleGAN \cite{image2stylegan} fails to produce semantically meaningful image on this task.
  }
  \label{appendix:fig:face-diffusion-analysis}
\end{figure}

\begin{figure}[t]
  \centering
  \includegraphics[width=1.0\linewidth]{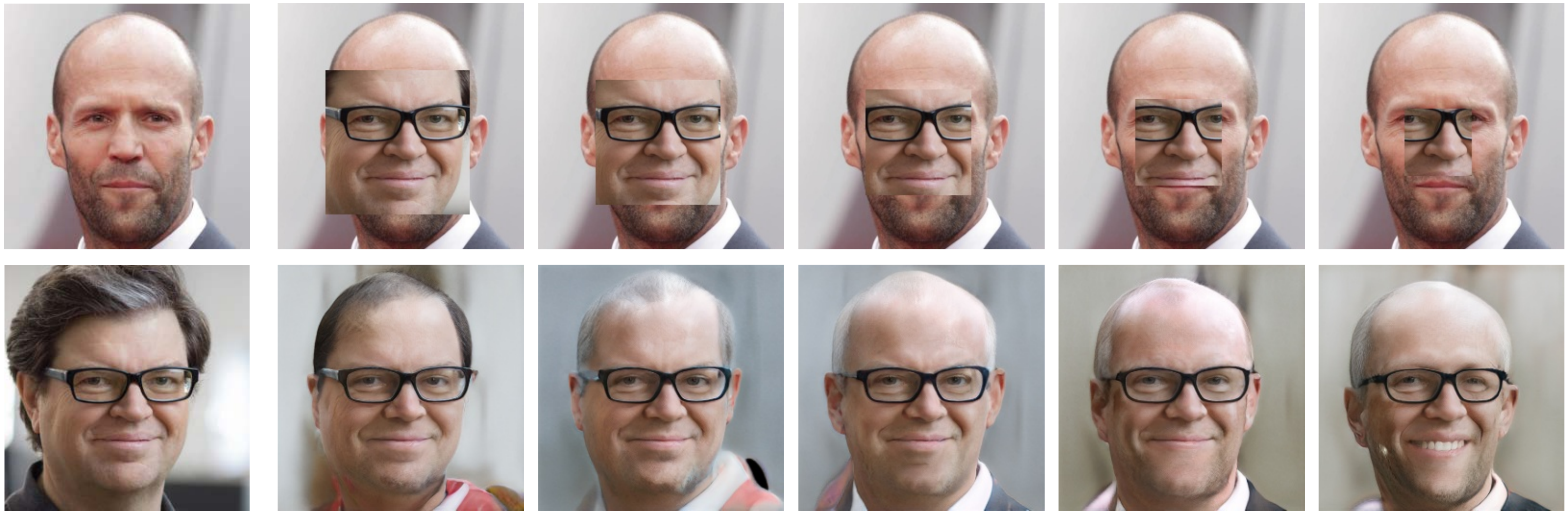}
  \caption{
    The effect of crop size on semantic diffusion.
    Top-left corner shows the context image while bottom-left corner shows the target image.
    Each remaining column correspond to a different crop size.
    Top row shows the direct copy-paste results, while bottom row shows the semantic diffusion results.
  }
  \label{appendix:fig:face-diffusion-size}
\end{figure}

\section{Semantic Diffusion}\label{appendix:diffusion}
In this part, we deeply analyze the semantic diffusion achieved by our \emph{in-domain} inversion.

\noindent\textbf{Implementation Details.}
Given a target-context image pair, we first crop the wanted part from the target image and then paste it onto the context image.
Then, we use our \emph{domain-guided} encoder to infer the latent code for the stitched image.
Due to the domain-alignment property of our encoder, the reconstruction from the code can already capture the semantics from both the target patch and its surroundings and further smooth the contents (see the fourth column in Fig.\ref{appendix:fig:face-diffusion-analysis}).
With this code as an initialization, we finally perform masked optimization by only using the target foreground region to compute the reconstruction loss.
In this way, we are able to not only diffuse the target image to any other context, but also keep the original style of the context image.

\noindent\textbf{Comparison.}
We show the intermediate results of the semantic diffusion process and compares our approach with the MSE-based inversion approach, Image2StyleGAN \cite{image2stylegan}.
The results are shown in Fig.\ref{appendix:fig:face-diffusion-analysis}, where we have three following observations:
(i) The output from our encoder always leads to the reconstruction of a meaningful face and keep most semantics of the inputs (\emph{e.g.}, gender and hair). That is because all codes produced by our encoder are \emph{in-domain}.
(ii) The masked optimization is able to preserve the identity information of the target face and further diffuse its style (\emph{e.g.}, skin color) to the surroundings, leading to seamless fusion. This step barely affects the context style (\emph{e.g.} hair style) inherited from the encoder initialization.
(iii) Image2StyleGAN fails to produce semantically meaningful faces (\emph{e.g.}, not smooth on the stitch boundary) in the diffusion task since they only focus on the reconstruction of pixel values but not semantics. By contrast, our \emph{in-domain} inversion achieves more satisfying results.

\noindent\textbf{Effect of Crop Size.}
We further studied the impact of the crop size on semantic diffusion.
As shown in Fig.\ref{appendix:fig:face-diffusion-size}, we can see that the larger the crop size is (\emph{i.e.} larger reference region from target face), the better the identity information is preserved.
For example, on the second column of Fig.\ref{appendix:fig:face-diffusion-size}, even hair is transmitted from the target image to the context image since the temples are included in the cropped patch.
On the last column, however, the diffused result is no long like the target face at all (\emph{e.g.}, the facial shape and mouth).
That is because, during the process of masked optimization, only the foreground patch is used as reference.
The surroundings will adaptively change starting from the encoder initialization.
Even so, thanks to the \emph{in-domain} property, our approach is still able to complete the entire eyeglasses and generate a smooth diffusion result.

\end{document}